\documentclass[lettersize,journal]{IEEEtran}

\usepackage{amsmath,amsfonts}
\usepackage{array}
\usepackage{textcomp}
\usepackage{stfloats}
\usepackage{url}
\usepackage{verbatim}
\usepackage{graphicx}
\usepackage{cite}
\usepackage{algorithm2e}
\usepackage{algpseudocode}
\RestyleAlgo{ruled}

\usepackage{graphics} 

\usepackage{xcolor}

\usepackage{subfigure}

\begin{document}

\title{Visual Localization and Mapping in Dynamic and Changing Environments}

\author{Joao Carlos Virgolino Soares$^{1}$, Vivian Suzano Medeiros$^{2}$, Gabriel Fischer Abati$^{3}$, Marcelo Becker$^{2}$, \\Glauco Caurin$^{2}$, Marcelo Gattass$^{4}$ and Marco Antonio Meggiolaro$^{3}$

\thanks{$^{1}$ J. Soares is with the Department of Mechanical Engineering, University of Illinois at Urbana-Champaign, 105 S Mathews Ave, Urbana, USA.
        {\tt\small virgolinosoares@gmail.com}} 
\thanks{$^{2}$V. S. Medeiros, M. Becker and G. Caurin are with the Department of Mechanical Engineering, University of São Paulo, São Carlos, SP - Brazil
        {\tt\small viviansuzano@usp.br} /
        {\tt\small becker@sc.usp.br} /
        {\tt\small gcaurin@sc.usp.br}}
\thanks{$^{3}$G. Abati and M. Meggiolaro are with the Department of Mechanical Engineering, Pontifical Catholic University of Rio de Janeiro, R. Marques de Sao Vicente 225, Gavea, Rio de Janeiro, RJ - Brazil
        {\tt\small fischerabati@gmail.com} /
        {\tt\small meggi@puc-rio.br}}%
\thanks{$^{4}$M. Gattass is with the Tecgraf Institute, Pontifical Catholic University of Rio de Janeiro, R. Marques de Sao Vicente 225, Gavea, Rio de Janeiro, RJ - Brazil
        {\tt\small mgattass@tecgraf.puc-rio.br}}%
        
}

\maketitle

\begin{abstract}
The real-world deployment of fully autonomous mobile robots depends on a robust SLAM (Simultaneous Localization and Mapping) system, capable of handling dynamic environments, where objects are moving in front of the robot, and changing environments, where objects are moved or replaced after the robot has already mapped the scene. This paper presents Changing-SLAM, a method for robust Visual SLAM in both dynamic and changing environments. This is achieved by using a Bayesian filter combined with a long-term data association algorithm. Also, it employs an efficient algorithm for dynamic keypoints filtering based on object detection that correctly identify features inside the bounding box that are not dynamic, preventing a depletion of features that could cause lost tracks. Furthermore, a new dataset was developed with RGB-D data especially designed for the evaluation of changing environments on an object level, called PUC-USP dataset. Six sequences were created using a mobile robot, an RGB-D camera and a motion capture system. The sequences were designed to capture different scenarios that could lead to a tracking failure or a map corruption. To the best of our knowledge, Changing-SLAM is the first Visual SLAM system that is robust to both dynamic and changing environments, not assuming a given camera pose or a known map, being also able to operate in real time. The proposed method was evaluated using benchmark datasets and compared with other state-of-the-art methods, proving to be highly accurate.
\end{abstract}


\section{Introduction}

Simultaneous Localization and Mapping (SLAM) is a fundamental problem in robotics, especially considering the challenges of real-world scenarios, such as the presence of dynamic objects and environments where objects can be constantly removed, added or moved to different locations.

Recent works have presented visual SLAM systems able to work in highly dynamic environments, such as DynaSLAM~\cite{dynaslam}, DS-SLAM~\cite{dsslam}, SaD-SLAM~\cite{sadslam}, and DOTMask~\cite{dotmask}. However, none of them are able to deal with other types of dynamic factors that can happen in a real environment, such as changing scenes, where objects are moved outside the field of view of the robot, after the scene had already been mapped. 

On the other hand, there are methods that deal with changing environments that consider the pose of the camera known, i.e., do not perform SLAM, but mapping with known poses~\cite{gomez2020ral}. Other methods~\cite{derner2021} perform localization in changing environments with a known map. DXSLAM~\cite{dxslam} has an increased robustness to changing environments, but fails in regular dynamic environments. In general, the SLAM methods found in the literature that are robust to both dynamic and changing environments only perform 2D SLAM using LiDAR fused with other sensors such as IMU and odometry~\cite{2dchanging2021}. 

In this context, this paper presents a real time visual SLAM system able to work with highly dynamic environments and changing environments. A combination of a robust object tracker and a filtering algorithm enables our visual SLAM system to perform well in highly dynamic environments containing moving objects. Also, the system maintains a semantic map of the environment, updating the belief about the poses of objects in time, making it also robust to changing environments.

\subsection{Contributions}

This work proposes Changing-SLAM, a real-time visual SLAM system based on ORB-SLAM3. To the best of our knowledge, this is the first methodology for Semantic SLAM in both dynamic and changing environments, with the following contributions: \\

\noindent
1. \textit{Robustness to highly dynamic environments}: The proposed method uses a robust keypoint classification algorithm that filters a priori dynamic objects and uses an Extended Kalman Filter to track movable objects in the scene. The problem of feature depletion caused by filtering features from the background in the bounding boxes is solved with a fast and reliable method, using statistical data of the depth in each bounding box, called feature repopulation. \\

\noindent
2. \textit{Robustness to changing environments}: The system uses the MapPoints derived from feature detection, combined with the output of an object detector to determine the 3D centroid of the objects in the scene, and create an object-level semantic map that maintains a belief about the pose of each mapped object. This results in a real-time 3D object detection using a semantic point clustering approach, without the need for instance or panoptic segmentation or an off-the-shelf 3D object detector. A robust long-term data association is also proposed, using the object's centroids. The state of the objects in the map is updated using a Bayesian filter. Different from the approaches found in the literature, the proposed method does not assume a known camera pose, nor a known map \textit{a priori}. \\

\noindent
3. \textit{The PUC-USP dataset for changing environments}: Public datasets are fundamental elements for the evolution of SLAM systems. In contrast to other datasets publicly available, this work presents a dataset especially designed for the evaluation of the robustness of visual SLAM methods in changing environments. The data is collected using an RGB-D camera attached to a mobile robot, while a motion capture system is used to generate the ground-truth. It consists of sequences recorded in an indoor environment showing simple and challenging situations: vanishing objects, objects that are moved to different positions and replaced objects. Evaluation metrics for the estimated trajectory show that the proposed sequences could lead to failure in pose estimation for SLAM systems not robust to changing environments. To our knowledge, this is the first dataset focused on this problem. \\

Changing-SLAM is evaluated using benchmark datasets such as the TUM dataset and the PUC-USP dataset for changing environments, and compared with several state-of-the-art methods, including ORB-SLAM3, DynaSLAM, SaD-SLAM, DOTMask, DXSLAM and Detect-SLAM, achieving better camera localization accuracy in both dynamic and changing environments.

\section{Related Work}

\IEEEpubidadjcol

\subsection{SLAM in Dynamic Environments}

Recent approaches to Visual SLAM focus on the highly dynamic environment problem. Some of them~\cite{detect-slam,sgc,xiao2019} rely on object detection combined with a dynamic feature point removal algorithm. The main problem with this approach is that it could lead to feature depletion and lost tracks in certain conditions, such as when a person is too close to the camera, or in a scene with many people. This happens because when the keypoints inside the bounding box are filtered, some keypoints that belong to the background are also filtered, as occurs in Dynamic-SLAM~\cite{xiao2019}. SGC-VSLAM~\cite{sgc} handled this problem by using optical flow and computing the fundamental matrix between two frames to decide which keypoints belong to objects, which is computationally demanding.  A solution to this problem is proposed in Section~\ref{sec:keypointclass}, with a robust and fast feature repopulation algorithm for object detectors.

Another approach to solve this problem is to use an instance segmentation framework to differentiate objects from the background, but the inference time of instance segmentation networks is very high. DynaSLAM~\cite{dynaslam} uses the Mask R-CNN~\cite{mask} instance segmentation framework to obtain the pixel-wise information of people in the scene, using it to filter a priori dynamic features. Despite its high accuracy and robustness, it cannot perform real-time due to the high computational requirement of the Mask R-CNN framework. Similarly, DP-SLAM \cite{dpslam} combines the semantic information of Mask R-CNN with a geometric approach based on epipolar geometry and probability propagation to classify dynamic keypoints. SaD-SLAM \cite{sadslam} combines depth information and Mask R-CNN instance segmentation to find dynamic features in the image. Each feature point is individually classified as static, dynamic, or static and movable. SaD-SLAM has a high accuracy, higher than DynaSLAM \cite{dynaslam} in some scenarios. Its main drawback is that the semantic segmentation is processed offline.

DOTMask~\cite{dotmask} uses instance segmentation to obtain the pixel-wise information of the objects in the image, and an Extended Kalman Filter to track these objects. Their aim was to provide a faster SLAM system in exchange of a lower accuracy, in comparison with DynaSLAM, for example. The main problem with this approach is that the use of instance segmentation makes it still too slow, and the accuracy is considerably lower than SaD-SLAM \cite{sadslam} or DynaSLAM \cite{dynaslam}. 

Ji et al.~\cite{ji2021} proposed a faster Semantic RGB-D SLAM method for dynamic environments extracting semantic information only from keyframes. Also, they combined K-Means with depth reprojection to identify unknown moving objects in the other frames. Despite achieving an accuracy comparable with DynaSLAM with less computational effort, their tracking thread runs at approximately 13 FPS.

Some works consider people as \textit{a priori} dynamic objects, such as \cite{dynaslam} and \cite{soares_jint}. The assumption of considering people as dynamic \textit{a priori} may seem strong, but in reality people are dynamic by nature and they eventually move. Mapping a scenario where most people are static for a long period is unrealistic. Furthermore, even if people in a scene are static, mapping them would lead to a future wrong loop closure when revisiting that scene after they have moved. One way to overcome this issue is to use features from static people only for tracking purposes, as done in SaD-SLAM \cite{sadslam}.

\subsection{Dataset for Visual SLAM in Changing Environments}

Datasets and benchmarks are very important for the advances of SLAM research, as they provide an accessible way for comparing multiple methodologies and evaluate them with clear criteria. There are several datasets for visual SLAM in the literature, each one focused on a different problem, with different types of raw data and ground-truth. 

The KITTI dataset~\cite{kitti} is used for the evaluation of several outdoor problems, including visual odometry, visual SLAM, multi-object tracking, segmentation, among others. It contains monocular, stereo and RGB-D data.

The TUM RGB-D dataset~\cite{benchmark} is one of the most used for evaluating visual SLAM systems. It has 39 sequences of static scenarios, scenes with dynamic objects, with low texture, among others. It uses two evaluation metrics: the absolute trajectory error, which is suited to evaluate SLAM systems, and the relative pose error, which is suited to evaluate visual odometry drift. The ground-truth was made using a motion capture system. Similar to the TUM dataset is the Bonn RGB-D~\cite{re-fusion}. It uses the same evaluation metrics from the TUM dataset, but with the focus on highly dynamic scenarios.

The previously mentioned datasets are not designed to deal with the changing environment problem, but rather focused on dynamic objects appearing in front of the camera. On the other hand, the OpenLORIS-Scene Dataset~\cite{openloris} was developed for real environments with several challenges that were not embraced by past datasets, such as changing environments, changing view point, and illumination. 

Zhao et al.~\cite{2dchanging2021} proposed a framework for lifelong localization and 2D mapping, and released a dataset with several sequences of indoor and outdoor changing environments, such as markets, parking lots and offices. The dataset contains 2D and 3D LIDAR, IMU and wheel encoder data. However, their dataset do not include RGB-D data. Furthermore, their ground-truth is not made using external measurement sensors, such as a motion capture system.

The majority of SLAM systems that deal with changing environments do not rely on publicly available datasets for performance evaluation. Most of them carry out their own experiments to record data sequences more suited to the changes they expect to handle~\cite{lee2014solution,Rosen2016TowardsLF,practical_persistence,gomez2020ral,efficient_longterm,derner2021,schmid2021}. These experiments demand the use of several sensors and a platform for data collection, which can be expensive and time-consuming. A publicly available dataset focused on changing environments would greatly contribute for the advance of this field.

\subsection{Visual SLAM in Changing Environments}

One of the situations usually not considered in methods for dynamic environments, such as~\cite{dynaslam,dsslam,sadslam}, is when a change happens after the robot has already mapped the scene. When it revisits the scene, some objects are in different locations, some are missing, and new objects may have appeared. This is often referred in the literature as SLAM in low dynamic environments \cite{gomez2020ral}\cite{walcott2012dynamic}, semi-static environments \cite{Rosen2016TowardsLF}, changing environments \cite{2dchanging2021}, or simply long-term mapping \cite{bore2018}.

The term ``changing environments'' was chosen as the more appropriate for the task, as ``low dynamic'' or ``semi-static'' can be used in the context of a scene with objects moving slowly in front of the camera, and ``long-term mapping'' emphasizes the scalability issue.

An early solution to this problem was proposed by Walcott-Bryant et al. \cite{walcott2012dynamic} in 2012. They proposed a method for planar indoor environments with robots using laser scanners. They proposed a dynamic pose-graph that could be edited, removing poses according to scan matching results.

Lee and Myung \cite{lee2014solution} showed through experiments that the wrong loop closures caused by a moved object could not be solved by pose-graph optimization techniques robust to outliers, such as Switchable Constraints \cite{sunderhauf2012switchable}, Max Mixtures \cite{maxmixtures} or Dynamic Covariance Scaling \cite{agarwal2013robust}. 

Rosen et al. \cite{Rosen2016TowardsLF} proposed a method to model environmental change of features over time, called feature persistence, using a recursive Bayesian estimator. Hashemifar and Dantu \cite{practical_persistence} extended Rosen's formulation, incorporating the persistence filter to ORB-SLAM and testing in a real environment.

Gomez et al. \cite{gomez2020ral} developed a method for dealing with changing environments on an object level. To create a 3D bounding box of an object, they use 2D object detection and point cloud to estimate the centroid position and object dimensions. They use a floodfill algorithm and the median of the 2\% smallest depths within the 2D bounding box to extrapolate the maximum and minimum depths of the object. Also, they create an object-based pose graph, connecting the robot poses and objects. The graph is updated computing the probability of finding the object in that location based on new measurements. The main drawback of their formulation is that the robot always revisits the same locations to update the object-graph. Thus, they do not perform SLAM, but mapping with known poses.

Zhao et al. \cite{2dchanging2021} proposed a framework for lifelong localization and 2D mapping, tracking the changes in the scene and maintaining an updated map accordingly through a technique called pose-graph refinement. Their method uses IMU, wheel encoders and LiDAR measurements. Lazaro et al. \cite{efficient_longterm} also proposed a method for changing environments using laser scans.

Derner et al. \cite{derner2021} proposed a method for visual localization in changing environments. Their method uses a previously built visual database, used to perform matching against query images to determine the pose of the robot.

Schmid et al.\cite{schmid2021} proposed a method for mapping in changing environments using panoptic segmentation to build and maintain volumetric maps during operation, receiving robot poses from an external estimator.

A survey of robust SLAM systems from 2021~\cite{arewethere} mentions DX-SLAM~\cite{dxslam} as the only example of a visual SLAM system designed for lifelong operations. DX-SLAM is a visual SLAM method that uses features from a deep convolutional neural network. Despite considerably improving robustness in changing environments, deep features alone did not improve robustness in dynamic environments. 

None of the mentioned methods deal with both dynamic and changing environments using cameras. The ones that deal with changing environments are either not robust to highly dynamic environments, or assume a known camera pose or a known map, i.e., do not perform SLAM.

\section{METHODOLOGY}


The semantic SLAM problem can be formulated as stated in Eq. (\ref{eq:semantic_slam}), where $\hat{X}$ is the set of estimated robot poses, $\hat{L}$ is a set of estimated landmark positions and semantic classes, and $Z$ is a set of sensor measurements \cite{doherty_icra2019}.

\begin{equation}
\hat{X},\hat{L} =    \underset{X,L}{\mathrm{argmax}} \thinspace p(X,L | Z)
\label{eq:semantic_slam}
\end{equation}

In the proposed approach, however, the semantic SLAM formulation is decoupled from the geometric one. An object belief map is created and updated acting as a pre-filter for the standard feature-based SLAM procedure, making the SLAM process occur in two levels. Within the high level (objects), a Bayesian filter is used to create a belief map about the poses of objects in the map. This belief map decides which features are used for the low-level step. Within the low level (points), the SLAM problem is solved using the feature-based methods proposed by ORB-SLAM3, adapted for dynamic environments. This approach results in a reliable tracking system, robust to changes in the map, and a semantic map in an object-level that can be used for other problems such as autonomous navigation.

\begin{figure*}[h!]
	\centering
	\includegraphics[width =0.75\linewidth]{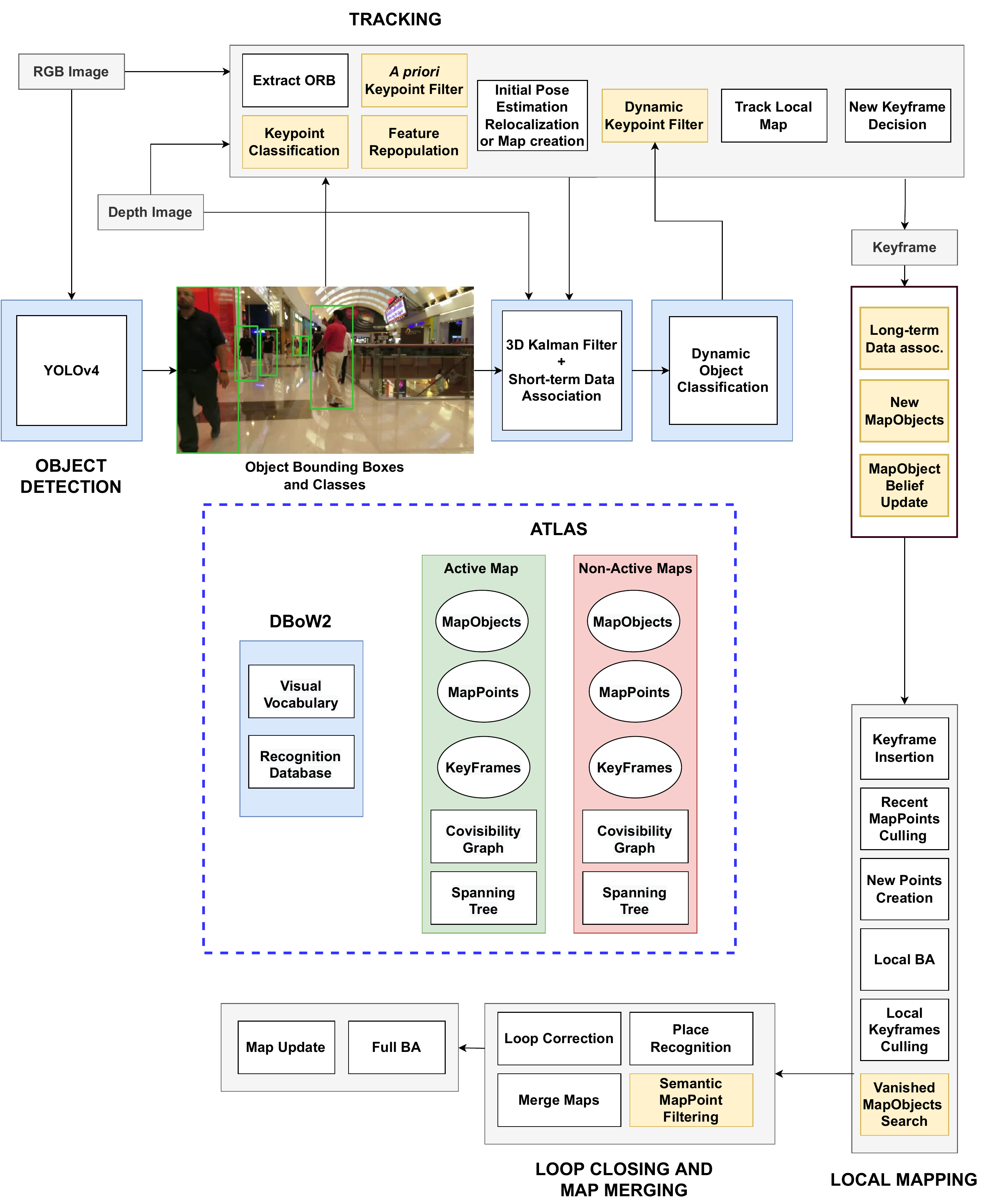}    
	\caption{Main components of Changing-SLAM}
	\label{fig:changing_pipeline}
\end{figure*}

Figure \ref{fig:changing_pipeline} shows the framework of the proposed methodology. The system is built on ORB-SLAM3, and it is composed of four threads running in parallel: Object Detection, Tracking, Local Mapping, and Loop Closing. Changing-SLAM modifies all three threads of ORB-SLAM3 and adds a new thread for object detection. ORB features are extracted from the RGB image in the tracking thread, and each associated keypoint is initially classified as dynamic, movable or static, according to the semantic information provided by the object detection thread. A feature repopulation algorithm is proposed to differentiate object features from background ones, and features from people are filtered \textit{a priori}.

The local mapping thread adds new keyframes and points to the map. The MapPoints, created from the detected and classified features, generate MapObjects with a semantic class and an unique ID. An extended Kalman filter is used to track the objects and predict their state, based on their ID. If an object has a velocity above a threshold, its keypoints are filtered from the image.

The loop closing thread and graph-optimization remain the same of ORB-SLAM3. The output is the pose of the camera frame by frame, processed with filtered sensor information, and the sparse map clear of outliers.

Changing-SLAM explores one of the main novelties of ORB-SLAM3, the Atlas framework \cite{orbslamatlas}. It is a multi-map system that can store a set of disconnected maps, and merge them when a loop is detected. This considerably improves the SLAM solution in scenarios with lost tracks. The proposed approach modifies Atlas to include storage and operations with MapObjects.

A long-term data association is also proposed to decide whether an object detected in the current frame is already in the map, or if it is a new object. The long-term data association is also responsible for updating the belief about the persistence of the objects in the environment. Finally, MapPoints associated to MapObjects with a low belief are filtered from the loop closing and graph-optimization steps.

\subsection{Keypoint Classification} \label{sec:keypointclass}

The keypoints detected in the tracking thread are initially classified into three categories: dynamic, movable or static. All keypoints belonging to people are classified as \textit{a priori} dynamic, keypoints that belong to objects are classified as movable, and the keypoints from the background are static.

\begin{figure}[ht]
    \centering
    \includegraphics[width = \linewidth]{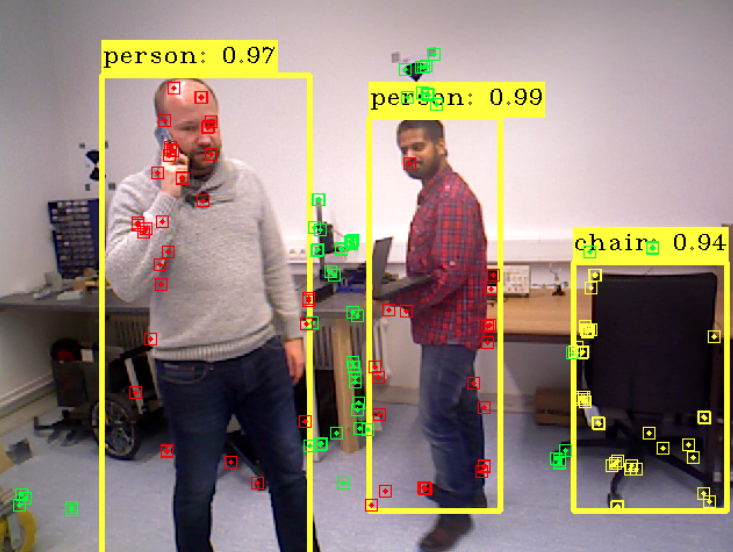}
    \caption{Initial Keypoint Classification. Dynamic keypoints are red, movable keypoints are yellow, and static keypoints are green.}
    \label{fig:keypointclass}
\end{figure}

Different from SaD-SLAM \cite{sadslam}, two keypoints that belong to the same object cannot have different classifications. This improves the speed of the process, because evaluating the dynamics of each individual keypoint is unfeasible in real time. Figure \ref{fig:keypointclass} shows an example of the initial keypoint classification being performed in a frame with two people and one chair.

Using object detection for feature removal can lead to a depletion of features, especially when there are many dynamic objects in the scene, or when a dynamic object occupies a large portion of the image. Instead of using computationally expensive methods based on epipolar geometry and RANSAC, this work presents an efficient method to correctly classify the features that belong to the bounding box but are not dynamic, called feature repopulation.

In each bounding box the median, mean, maximum and minimum pixel depth values are extracted, as well as the standard deviation. With this information, together with the IoU matrix, it is possible to evaluate whether the detected object is being occluded, and to differentiate the object from the background.

\begin{figure}[h!] 
  \subfigure[   \hspace{0.05cm}     Without feature repopulation]{%
    \includegraphics[width=.459\linewidth]{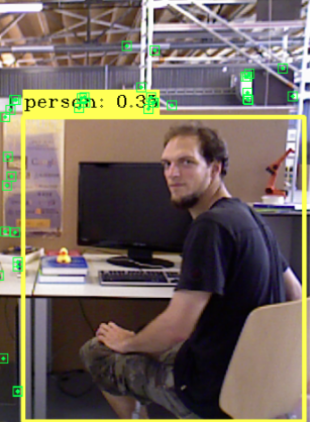} \label{fig:old_filtering_dyn}
  } 
  \subfigure[   \hspace{0.05cm}     With feature repopulation]{%
    \includegraphics[width=.485\linewidth]{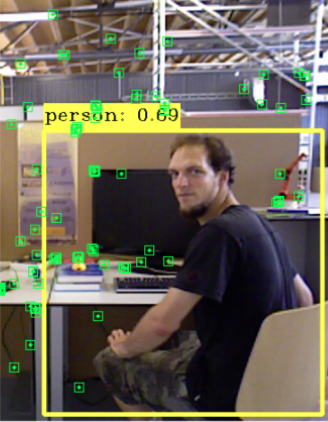} \label{fig:new_filtering_dyn}
  } 
  \caption{\textit{A priori} people keypoint filtering proposed in this methodology. The figure shows the effect of the feature repopulation technique that keeps keypoints that belong to the background but are located inside the bounding box.}
\end{figure}

To correctly classify the keypoints, it is necessary to consider a few possible conditions. If the object is not being occluded, then the classification is straightforward. Keypoints with a depth greater than the minimum depth plus a threshold value are considered belonging to the background. The threshold depends on the class of the object. If the object is being partially occluded by another known detected object, the depths inside the overlapping area are not considered. This occlusion is evaluated calculating the IoU between the bounding boxes of the frame. The main problem arises when a non-labeled or non-detectable object is occluding the target object. This problem is identified when the standard deviation of the depths is too high, or if the median depth is higher than the depth of the center of the bounding box. If this happens, it means that the center of the bounding box is being occluded. In this last scenario, the keypoints of this object are not considered. Figures \ref{fig:old_filtering_dyn} and \ref{fig:new_filtering_dyn} show the result of the feature repopulation technique. In Fig. \ref{fig:old_filtering_dyn}, all keypoints inside the bounding box are filtered out. However, in Fig. \ref{fig:new_filtering_dyn}, just the keypoints that belong to people are filtered out, while the keypoints from the background are kept. 

Algorithm \ref{alg:keypointclass} details the process of keypoint classification, feature repopulation, and keypoint filtering. The number of the class associated with each keypoint is the same number of that class in the COCO dataset \cite{coco}. For example, ``person'' is $0$, ``tvmonitor'' is $62$, and ``chair'' is $56$. When a keypoint has its class set as $-1$, it means that the keypoint belongs to the background.

Algorithm \ref{alg:keypointclass} shows the process for people, but the same procedure is done with movable objects. However, instead of being filtered \textit{a priori}, keypoints that belong to movable objects receive a class number corresponding to the object class.

\begin{algorithm}
\SetAlgoLined
\KwData{Current frame $F_k$, Last frame $F_{k-1}$, bounding box list $D^{F_k}$(x,y,w,h, class), keypoint list $p^{F_k}$, DOC} 

firstloop = true;\\
nbox = 0;\\
new\_keypoints;\\

\If{$D^{F_k}$ size $>$ 0}{
    \For{$p_i$ in $p^{F_k}$}{
        bool\_key = false;

        \While{nbox $<$ $D^{F_k}$ size}{
            
            \If{is the first loop}{
                \text{Get bounding box depth statistics}
            }
            \For{bb in $D^{F_k}$}{
                \text{Check occlusion between $D^{F_k}[nbox]$} \\ 
                \text{and $D^{F_k}[bb]$}
            }
            \If{$D^{F_k}$(class) is person}{
                \If{$p_i$ is inside $D^{F_k}[nbox]$}{
                    bool\_key = true;\\
                    \If{no occlusion by a labeled object}{
                        \If{no occlusion by unlabeled object}{
                            \If{$p_i$ depth $>$ mindepth + depth threshold}{
                                bool\_key = false;\\
                                keypoint class = -1; \\
                            }
                            \Else{
                                keypoint class = $D^{F_k}$(class);
                            }
                        
                        }
                        \Else{
                            bool\_key = false;\\
                            keypoint class = -1; \\
                            
                        }
                    }
                    \Else{
                        disregard the occluded part and perform the steps of lines 18 to 30;
                    
                    }

                }
            }

            $nbox++$;
        }
        firstloop = false;\\
        \If{bool\_key is false}{
            new\_keypoints receive $p_i$
        }
    }
    $p^{F_k}$ = new\_keypoints;
}

 \caption{Keypoint classification and filtering}
  \label{alg:keypointclass}
\end{algorithm}

\subsection{Map Objects}

MapPoints are one of the key elements in ORB-SLAM3 \cite{orbslam3}. They are created from the ORB features detected in each frame, and used for all tasks, including camera tracking, local optimization, and loop closure. Finally, they constitute the map representation. 

A MapObject is an element created to represent a group of MapPoints that belong to same entity, with a common body and class. In the proposed methodology, besides all other pieces of information needed for the ORB-SLAM3 framework, MapPoints store a 3D position $X_{w,i}$ in the world coordinates, the class, and the ID of the MapObject associated with the MapPoint. When the MapPoint is created, it receives the class and ID of the MapObject associated with the bounding box where the keypoint was located.  Each MapObject stores the first bounding box, the current bounding box, class, list of associated MapPoints, the 3D position in world coordinates, the unique global ID, the belief of persistence, and the 3D bounding box dimensions.

The belief of persistence is the numerical value that will determine whether the MapObject, and its associated MapPoints, will remain active in the map. Details of its initialization and update are given in Sections \ref{sec:objectpersistence} and \ref{sec:longtermda}. 

The 3D position is obtained by computing the centroid of the associated MapPoints. The maximum object dimensions are obtained by computing the median of the $5\%$ maximum and minimum x, y and z coordinates of the associated MapPoints.

\subsection{Short-term Data Association}
\label{sec:short-term_dataassoc}

The short-term data association evaluates if the new bounding boxes detected in the current frame correspond to the bounding boxes of the MapObjects present in the last frame. This is done using the intersection over union (IoU).

In each new frame, for every bounding box the IoU is computed with all detections of the same class in the previous frame. If no matches are found, a new MapObject instance is created. This process is detailed in Algorithm \ref{alg:shortterm}. When a new KeyFrame is created, the system checks whether a tracked MapObject has new associated MapPoints. If it has, then its pose is updated.


\begin{algorithm}
\SetAlgoLined
\KwData{Frame $F_k$, bounding box list $D^{F_k}$(x,y,w,h, class), list of last frame objects lastFrameObjs}

\For{det in $D^{F_k}$}{
    associationfound = false;

    \For{lfo in lastFrameObjs}{
        \If{lastFrameObjs not NULL}{
            iou = GetIOU(det, bounding box of lastFrameObjs(lfo));\\
            IOU\_Matrix(det)(lfo) = iou;\\
            \If{current\_class == class of lastFrameObjs(lfo)}{
                \If{IOU\_Matrix(det)(lfo) $>$ IOU\_Threshold}{
                    \text{associationfound} = true; \\
                    lastObjID = lfo;\\
                    break;\\
                }
            }
        }
    }
    \If{associationfound}{
        currentMapObjects receive lastFrameObjs(lastObjID);\\
    }
    \Else{
        Create new MapObject; \\
        currentMapObjects receive new MapObject;\\
    }    
}

 \caption{Short-term Data Association}
 \label{alg:shortterm}
\end{algorithm}

\subsection{Object Tracking}

The position of a MapObject is given by the 3D position of its centroid. The position of the centroid is obtained through the position of each associated MapPoint. With the short-term data association algorithm, presented in Section \ref{sec:short-term_dataassoc}, each new detection is matched with all MapObjects from the last frame, which allows the possibility to track objects over time. In order to filter dynamic keypoints, it is necessary to infer if the tracked objects are moving or not.

The Extended Kalman Filter (EKF) is a non-linear state estimator that considers motion and observations corrupted by a zero mean Gaussian noise. The objective of the EKF is to obtain the best estimate of $\mathbf{x}$ given the measurements $\mathbf{z}$. An EKF is used to track each MapObject, in order to predict which one is moving, and it is initialized for each new MapObject. The state of the object is defined as its 3D position and velocity, as stated by

\begin{equation}
    \mathbf{x} = [x \thinspace \thinspace y \thinspace \thinspace z \thinspace \thinspace \dot{x} \thinspace \thinspace \dot{y} \thinspace \thinspace \dot{z}]^{T}
    \label{eq:objstate}
\end{equation}

The prediction step at frame $k$ is given by

\begin{equation}
    \mathbf{\hat{x}}_{k|k-1} = \mathbf{F}\mathbf{\hat{x}}_{k-1|k-1} 
\end{equation}

\noindent where $k$ is the current frame, $k-1$ is the last frame, and $\mathbf{F}$ is given by 

\begin{equation}
\bf{F} = 
\begin{bmatrix}
    \mathbf{I}_{3} & \Delta t \mathbf{I}_{3} \\
    \mathbf{0}_{3} & \mathbf{I}_{3}  
    \end{bmatrix} 
    \label{eq:F_ekf}
\end{equation}

\noindent where $\Delta t$ is the time between predictions, $\mathbf{I_3}$ is a 3x3 identity matrix, and $\mathbf{0_3}$ is a 3x3 matrix with zeros. The state covariance is estimated as stated by

\begin{equation}
    \mathbf{P}_{k|k-1} = \mathbf{F}\mathbf{P}_{k-1|k-1}\mathbf{F}^{T} + \mathbf{Q} 
    \label{eq:statecov}
\end{equation}

 \noindent where $\mathbf{P}_0 = \mathbf{I}$. The measurement update is given by Eq. (\ref{eq:measurement_update}), and the innovation covariance by Eq. (\ref{eq:innovation}).

\begin{equation}
    \mathbf{\tilde{y}}_k = \mathbf{z}_k - \mathbf{H}_k\mathbf{\hat{x}}_{k|k-1}
    \label{eq:measurement_update}
\end{equation}

\begin{equation}
    \mathbf{S}_k = \mathbf{H}_k \mathbf{P}_{k|k-1}\mathbf{H}_k^{T} + \mathbf{R}
    \label{eq:innovation}
\end{equation}


\noindent where $\mathbf{z}_k$ correspond to the measured coordinates of the centroid. The $\mathbf{H}_k$ matrix transforms the predicted state from the world frame to the camera frame. $\mathbf{Q}$ and $\mathbf{R}$ are the process and observation covariance matrices, respectively. Finally, the Kalman gain is stated by 

\begin{equation}
    \mathbf{K}_k = \mathbf{P}_{k|k-1} \mathbf{H}_k^{T}\mathbf{S}_k^{-1}
    \label{eq:kalmangain}
\end{equation}

The updated state estimate and covariance are given by Eqs. (\ref{eq:stateestimate}) and (\ref{eq:stateestimate2}).

\begin{equation}
    \mathbf{\hat{x}}_{k|k} = \mathbf{\hat{x}}_{k|k-1} + \mathbf{K}_k \mathbf{\tilde{y}}_k
    \label{eq:stateestimate}
\end{equation}

\begin{equation}
    \mathbf{P}_{k|k} = (\mathbf{I}_{6} - \mathbf{K}_k\mathbf{H}_k)\mathbf{P}_{k|k-1}
    \label{eq:stateestimate2}
\end{equation}


\subsection{Dynamic Object Classification}

The output of the object tracking is the state of each tracked MapObject. The dynamic object classification module outputs a Boolean result, establishing if the object is moving or not in that particular frame, based on a threshold, called ``DOC threshold''.

Keypoints belonging to new objects are filtered \textit{a priori}. If the object tracker establishes that the object is static, the keypoints are used for feature tracking. Keypoints belonging to moving objects are classified with the same procedure as the one described in Algorithm \ref{alg:keypointclass}.

\subsection{MapObject Persistence}
\label{sec:objectpersistence}

This work proposes a recursive Bayes' filter to estimate the belief about the MapObjects' persistence in the map. When a MapObject is created, the belief is set to $0.5$.

H is a discrete random variable that represents persistence of a given MapObject initialized at a certain 3D position. It can assume the values 0 or 1, i.e., either the object is not there or it is there. The belief about the persistence of a given MapObject is stated by

\begin{equation}
    bel(H) = \eta \thinspace p(Y | H) p(H)
    \label{eq:belief_H}
\end{equation}

\noindent where $\eta$ is a normalization factor given by 

\begin{equation}
    \eta = \frac{1}{bel(H=1) + bel(H=0)}
    \label{eq:eta_belief}
\end{equation}

\begin{equation}
    bel(H = 1) = \eta \thinspace p(Y | H=1) p(H=1)
\end{equation}

\begin{equation}
    bel(H = 0) = \eta \thinspace p(Y | H=0) p(H=0)
    \label{eq:eta_belief2}
\end{equation}

For each iteration, the prior $p(H)$ is the last belief. The likelihood $p(Y|H)$ is measured using the distance between the centroids, the measured and the one in the map. If an object is not detected at the place it was previously seen, its belief is lowered. The belief update of each MapObject is performed during the long-term data association, which is explained in the next section.

\subsection{Long-term Data Association}
\label{sec:longtermda}

The long-term data association evaluates if the MapObjects created in the current frame correspond to existing objects in the map. This process is detailed in Algorithm \ref{alg:longterm}. First, the centroid of a MapObject candidate is computed using its associated MapPoints. Then, the Euclidean norm between the candidate's pose and close MapObjects with the same class is computed. If they are close enough, they are merged and their MapPoints are combined. Also, the belief of the object in the Map is updated accordingly.

If the belief of a MapObject is below a threshold (BeliefThreshold), then all its associated MapPoints are marked as inactive and cannot be used for tracking, mapping or loop closure. 
As all MapObjects start with a 0.5 belief, initially all MapPoints that belong to MapObjects are inactive. With this approach, objects that are moved or vanish from the map cannot interfere in the mapping process. As an object continues to be seen, its belief grows and, eventually, it is added to the map and its MapPoints become active.

\begin{algorithm}
\SetAlgoLined
\KwData{Current frame $F_k$, last frame $F_{k-1}$, list of MapObject candidates $newMO$, current Map, current keyframe $KF_k$, last keyframe ${KF}_{k-1}$, list of MapObjects in the current Map $ObjInMap$}

    \For{Obj in $newMO$}{
        \If{number of MapPoints in Obj  $>$ 0}{
            \If{Obj is new}{
                Compute centroid of Obj;\\
                merged = false;\\
                Get Obj pose $ObjPose$;\\
                \For{OiM in $ObjInMap$}{
                    Get OiM pose;\\
                    \If{class of Obj == class of OiM}{
                        dist = euclidean norm (Obj pose, OiM pose); \\
                        \If{$dist < ltdaThreshold $}{
                            merge objects (Obj, OiM);\\
                            merged = true;\\
                            \If{OiM was not saw in the past N Keyframes}{
                                Update Belief of OiM;\\
                                OiM last saw in $KF_k$;\\
                                OiMMP = Get all MapPoints from OiM;\\
                                \If{Belief $<$ BeliefThreshold}{
                                    OiMMP inactive;\\
                                }
                                \Else{
                                    OiMMP active;\\
                                }
                            }
                            \Else{
                                OiM last saw in $KF_k$;\\
                            }
                        
                        }
                    }
                }
                \If{merged is false}{
                    Add Obj in the Map;\\
                }
                
            }
            \ElseIf{Obj is not in $ObjInMap$}{
                perform steps of lines 4 to 33;\\
            }
            \Else{
                Compute centroid of Obj;\\
                Update Obj in map;\\
            }
            
            
        }
    
    }
    
     \caption{Long-term Data Association }
 \label{alg:longterm}
\end{algorithm}

\subsection{Semantic Map}

\begin{figure}[ht]
    \centering
    \includegraphics[width=\linewidth]{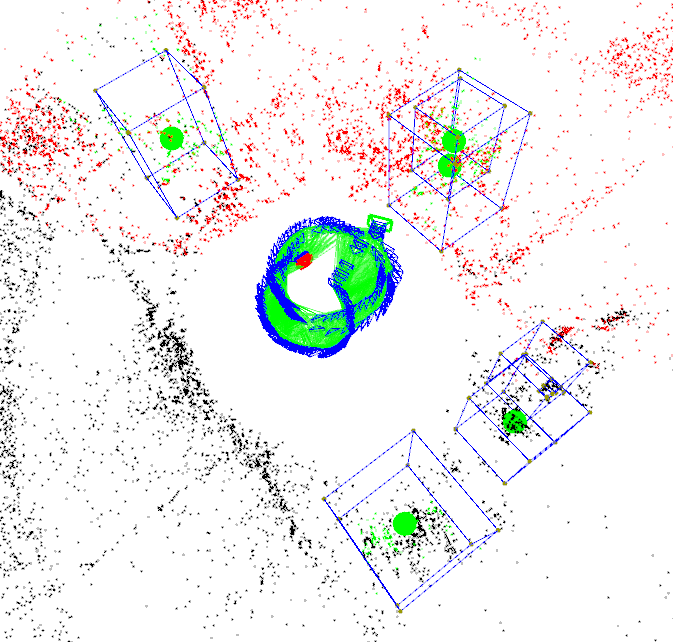}
    \caption{Semantic Map generated by Changing-SLAM}
    \label{fig:semanticmap}
\end{figure}

The final output of Changing-SLAM is the complete camera trajectory and a metric-semantic map. The metric map is composed of active MapPoints. The semantic map is composed of MapObjects with their respective centroids and 3D bounding boxes, as shown in Fig. \ref{fig:semanticmap}. Only objects with a high belief are active in the map. This information can be very useful for autonomous navigation.

\section{PUC-USP Dataset} \label{sec: puc-usp-dataset}

Experiments were performed in order to test the robustness of the methodology in changing environments and recorded as a dataset. It is composed of 6 sequences recorded in an indoor environment. Each sequence contains color and depth images captured by an Intel RealSense camera, and a ground-truth trajectory obtained using a motion capture system. The sequences were designed to capture different scenarios that could lead to a tracking failure or a map corruption. All data is publicly available.

\begin{figure}[h!]
	\centering
	\includegraphics[width=\linewidth]{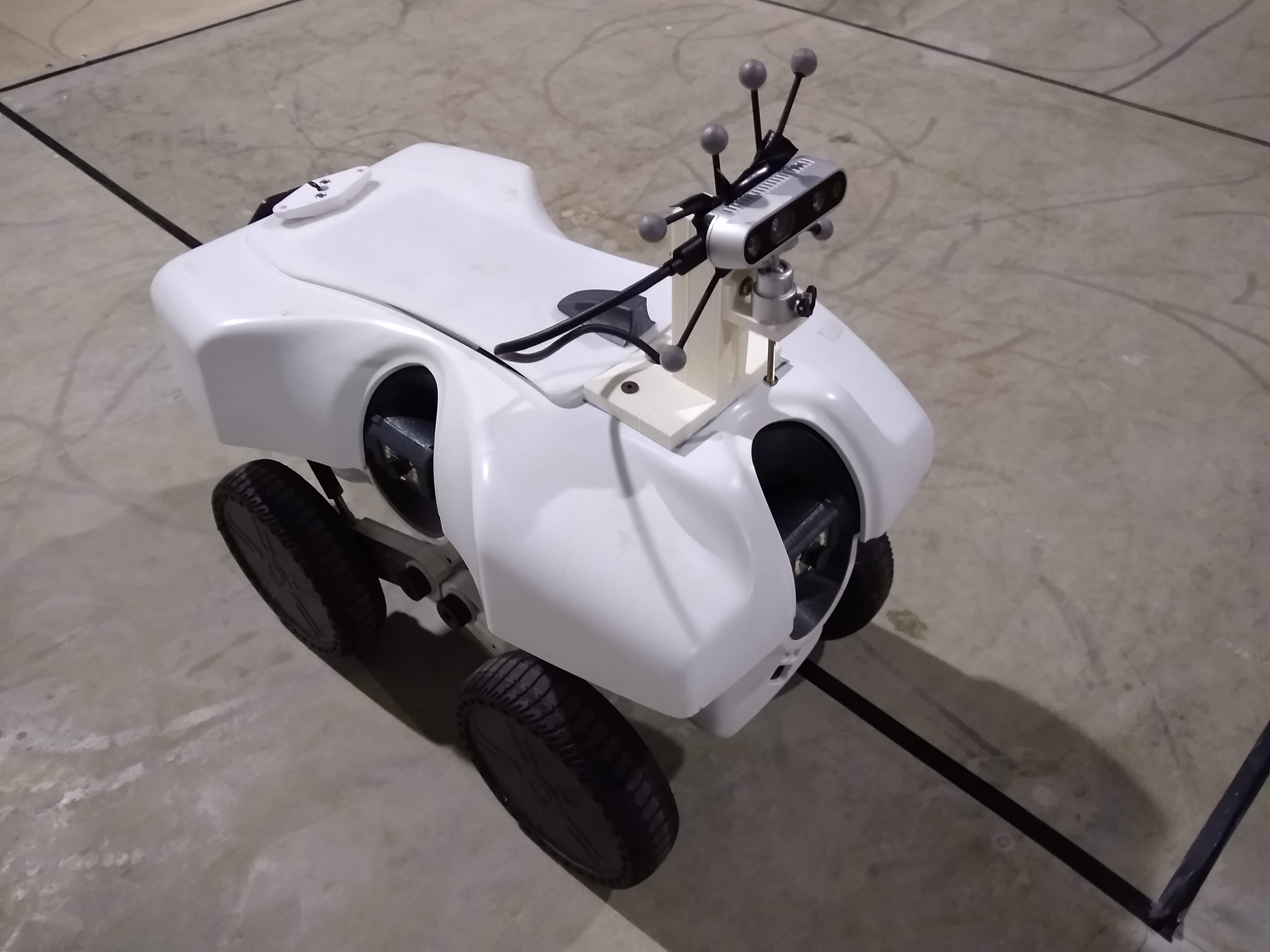} 
	\caption{TerraSentia Robot}
	\label{fig:terrasentia}
\end{figure}

The robot used in the experiments is shown in Figure \ref{fig:terrasentia}. The TerraSentia robot, developed by EarthSense \cite{earthsense}, is a mobile robot with a size of 0.32x0.54x0.4 meters and equipped with four active wheels for skid-steer locomotion. All data acquisition, processing, and locomotion control are performed using an Intel i7 NUC in combination with a RaspberryPi 3B board. The ground-truth was generated using a motion capture system, consisting of seven OptiTrack Flex13 cameras. The robot has an Intel RealSense D435i camera with reflective markers used by the OptiTrack system for computing the ground-truth trajectory. 

All data, including the RGB and depth images, is provided in the same format used in the TUM Dataset. The motion capture system was calibrated using the software provided by OptiTrack, and achieved a global precision of $4mm$ after the calibration procedure. The Network Time Protocol (NTP) was used to assure timestamp synchronization between the motion capture system and the camera.

The proposed dataset uses the Absolute Trajectory Error (ATE) for ground-truth evaluation, which is the same evaluation metrics from the TUM Dataset \cite{benchmark}. It is used to evaluate the global consistency of the estimated trajectory, comparing the absolute distances between the translational components of the estimated and ground-truth trajectories. Equation (\ref{eq:ATE}) shows the computation of the ATE at a time step $i$. 

\begin{equation}
    ATE_i = E_{i}^{-1}T G_i
    \label{eq:ATE}
\end{equation}

\noindent where $E$ represents the estimated trajectory, $G$ is the ground truth, and $T$ describes the transformation that aligns the two trajectories. For a sequence of $N$ poses, the RMSE of ATE is given by

 \begin{equation}
    RMSE(ATE_{1:N}) = \sqrt{\frac{1}{N} \sum_{i=1}^{N}||trans(ATE_i)||^2}
    \label{eq:RMSE_ATE}
\end{equation}

\noindent where $trans(ATE_i)$ correspond to the translational components of $ATE_i$.

Several objects from the COCO Dataset \cite{coco} were used in the sequences, such as a teddy bear, umbrellas, chairs and monitors. These objects were chosen because several semantic detectors use COCO for training, such as YOLOv3 \cite{yolov3}, YOLOv4 \cite{yolov4}, and Mask R-CNN \cite{mask}.

Six sequences were recorded in this dataset: \textit{Static}, \textit{OneChair}, \textit{Vanishing}, \textit{Changing}, \textit{Shift}, and \textit{Replacing}. All sequences were recorded with the camera fixed on the robot, with the robot moving in the workspace with flat terrain. Despite being fixed and located on a wheeled mobile robot, the motion of the Intel RealSense camera is not entirely restricted to a plane due to irregularities on the floor as well as robot vibration.

Table~\ref{tab:pucusp_sequences} shows the statistics over the six recorded sequences, containing the duration, total length, average translational velocity and average rotational velocity. The actual velocity of the robot was higher, because the robot eventually needed to stand still for some moments so the objects could be moved in the scene out of its field of view.

\begin{table}[h!]
\caption{Information about each sequence}
\label{tab:pucusp_sequences}
\begin{center}
\vspace{-4mm}
\renewcommand{\arraystretch}{1.2}
\setlength{\tabcolsep}{1.0pt}
\begin{tabular}{|c|c|c|c|c|c|}
\hline
Sequence & Duration [s] & Length [m] & Avg. T. Vel. [m/s] & Avg. R. Vel. [deg/s]\\
\hline
\textit{Static} & 109.73 & 9.09 & 0.0828 & 27.08\\
\hline
\textit{OneChair} & 294.83  & 18.78  & 0.0637  & 14.07 \\
\hline
\textit{Vanishing} & 277.53 & 23.17 & 0.0833  & 24.44\\
\hline
\textit{Changing} & 195.46  & 13.89 & 0.071  & 25.36\\
\hline
\textit{Shift} & 366.36  & 25.19 & 0.0687  & 16.26 \\
\hline
\textit{Replacing} & 174.53  & 15.49 & 0.0887  & 46.55\\
\hline
\end{tabular}
\end{center}
\end{table}

The \textit{Static} sequence is a baseline for the evaluations. No objects were moved in this sequence. The robot just wanders within a static scene. The \textit{Vanishing} sequence is suitable to evaluate the ability of SLAM systems to eliminate vanished objects in the map. It starts with the robot facing a chair with a teddy bear. As the robot wanders, some objects are moved and others are removed from the scene, until there are no more objects in the scene. Even though the missing objects would not cause a track failure or a wrong loop closure in a SLAM system not robust to changing environments, they would be present in the final map. This would interfere on a path planning algorithm that uses this map, for example.

\begin{figure}[t!]
	\centering
	\includegraphics[width=0.8\linewidth]{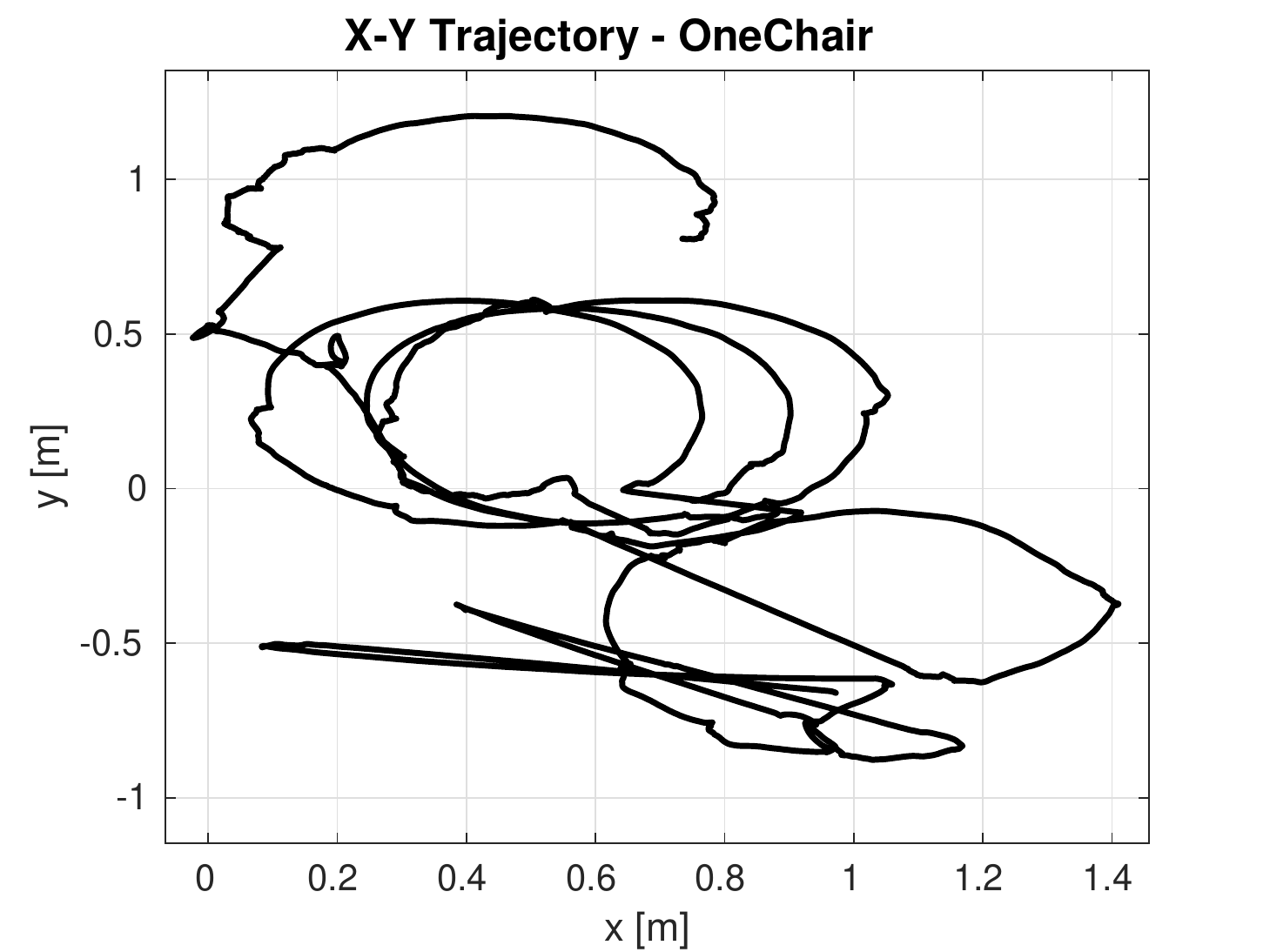}    
    \caption{Ground-truth trajectory of the \textit{OneChair} sequence}
 	\label{fig:onechair_traj}
\end{figure}

\begin{figure*}[h!] 
  \subfigure[        Scene 1]{%
    \includegraphics[width=.32\linewidth]{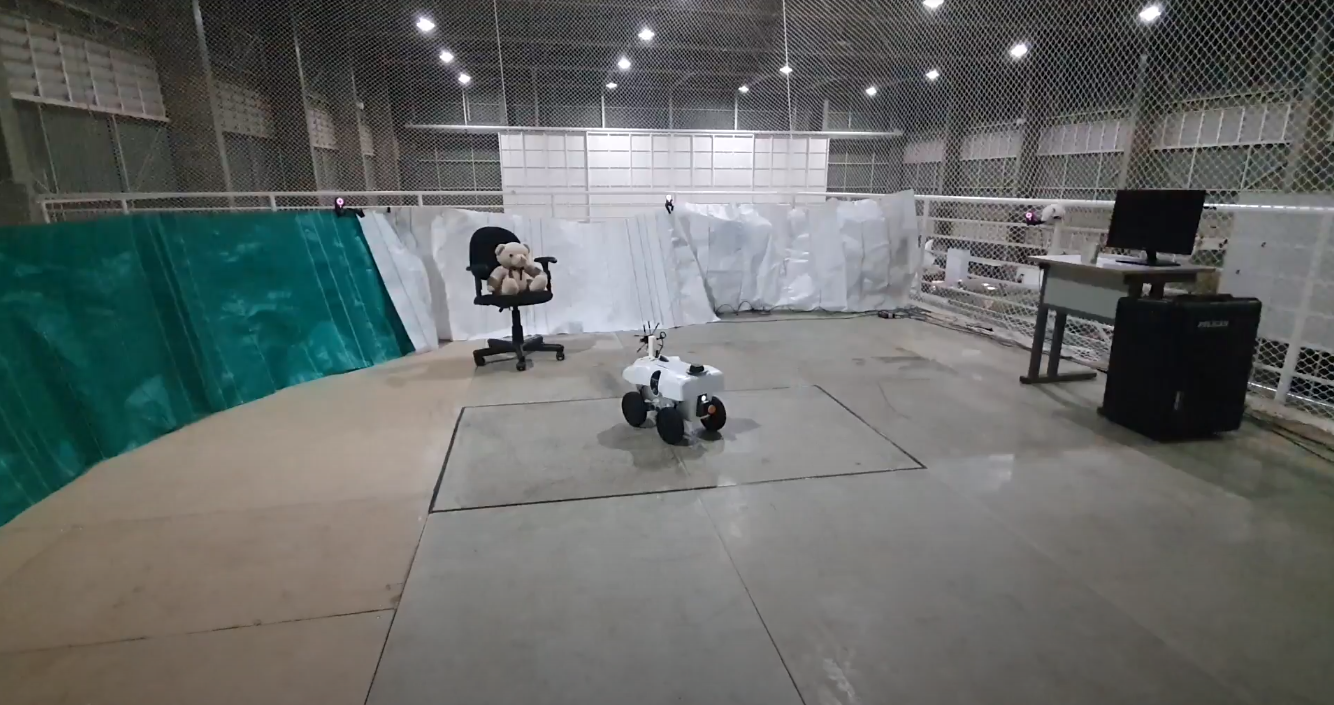} \label{fig:scene1_onechair}
  } 
  \subfigure[       Scene 2]{%
    \includegraphics[width=.32\linewidth]{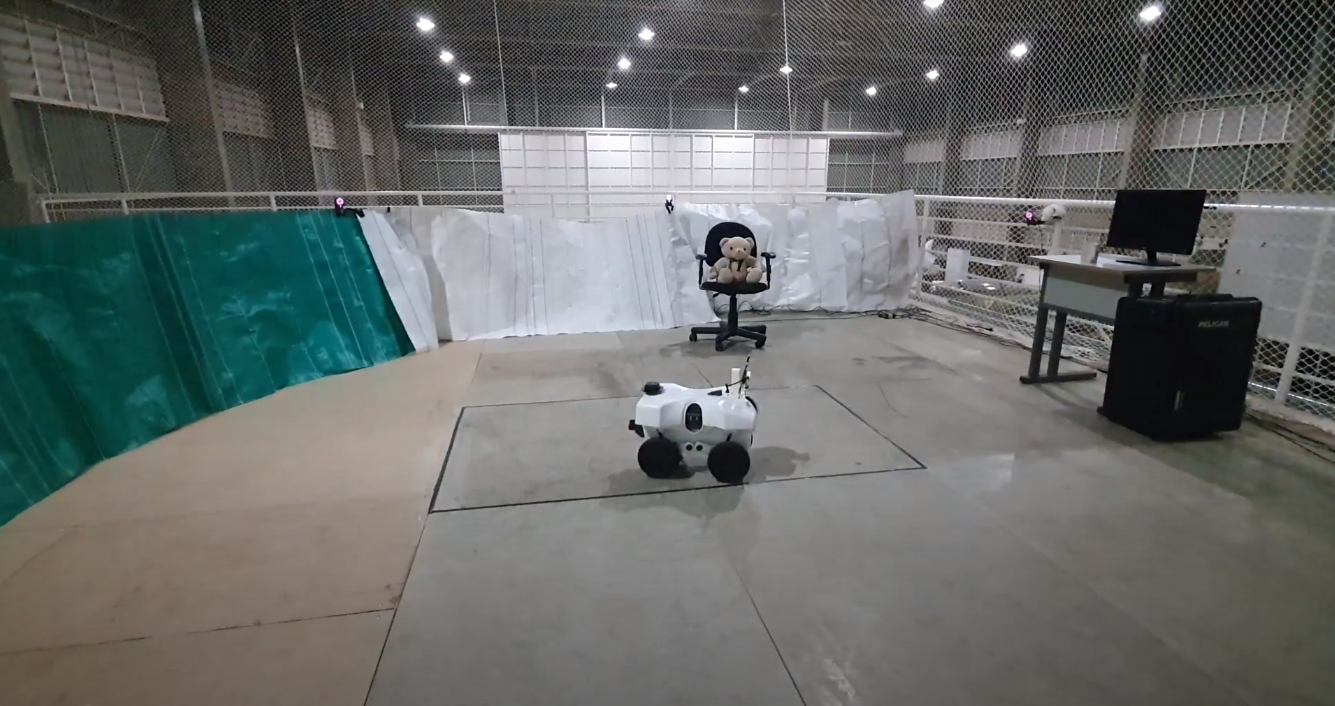} \label{fig:scene2_onechair}
  } 
  \subfigure[        Scene 3]{%
    \includegraphics[width=.32\linewidth]{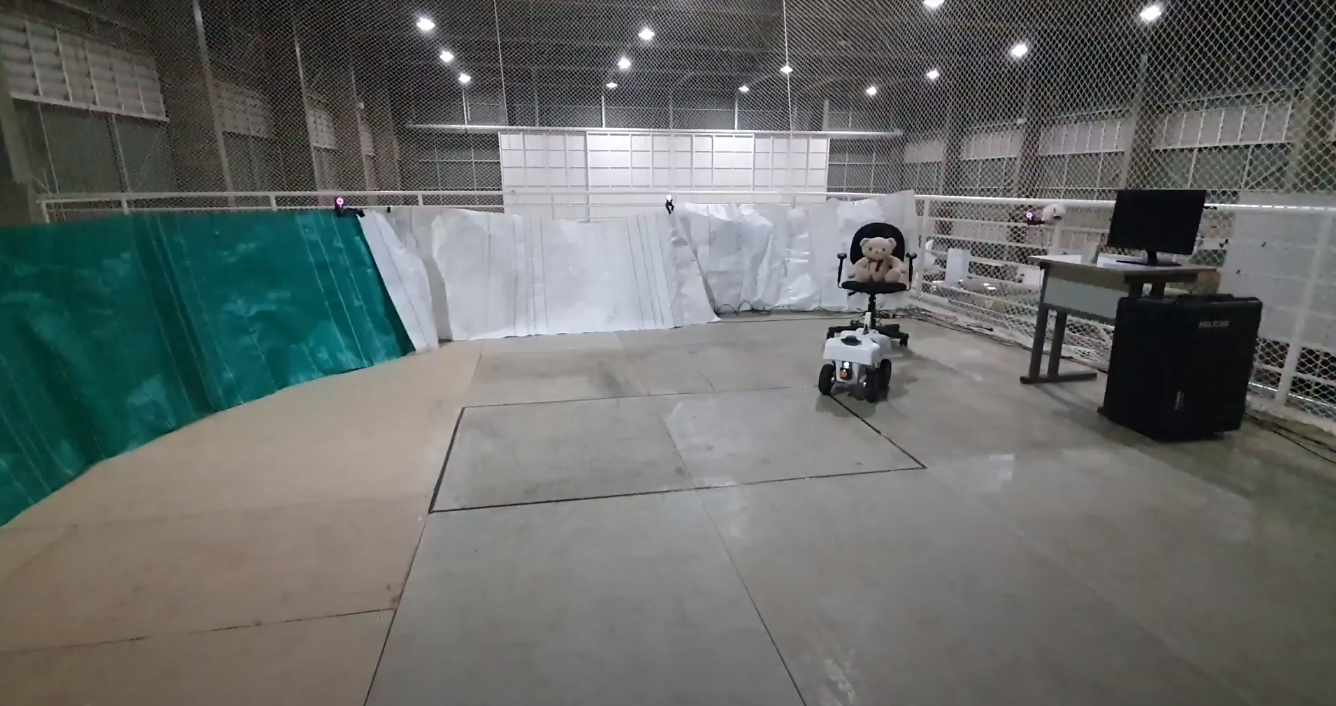}\label{fig:scene3_onechair}
  } 
  \caption{Selected scenes from the \textit{OneChair} sequence}
\end{figure*}

The \textit{OneChair} sequence is suitable to evaluate the ability of SLAM systems to avoid wrong loop closures caused by moved objects. Figure~\ref{fig:onechair_traj} shows the ground-truth trajectory of the \textit{OneChair} sequence. It starts with the robot facing a chair, as shown in Fig.~\ref{fig:scene1_onechair}. As the robot moves, the chair is also moved outside of the robot's field of view. When the robot sees the chair again, as shown in Fig. \ref{fig:scene3_onechair}, the chair is in a different position, which can cause a wrong loop detection.

\begin{figure*}[h!] 
  \subfigure[        Scene 1]{%
    \includegraphics[width=.32\linewidth]{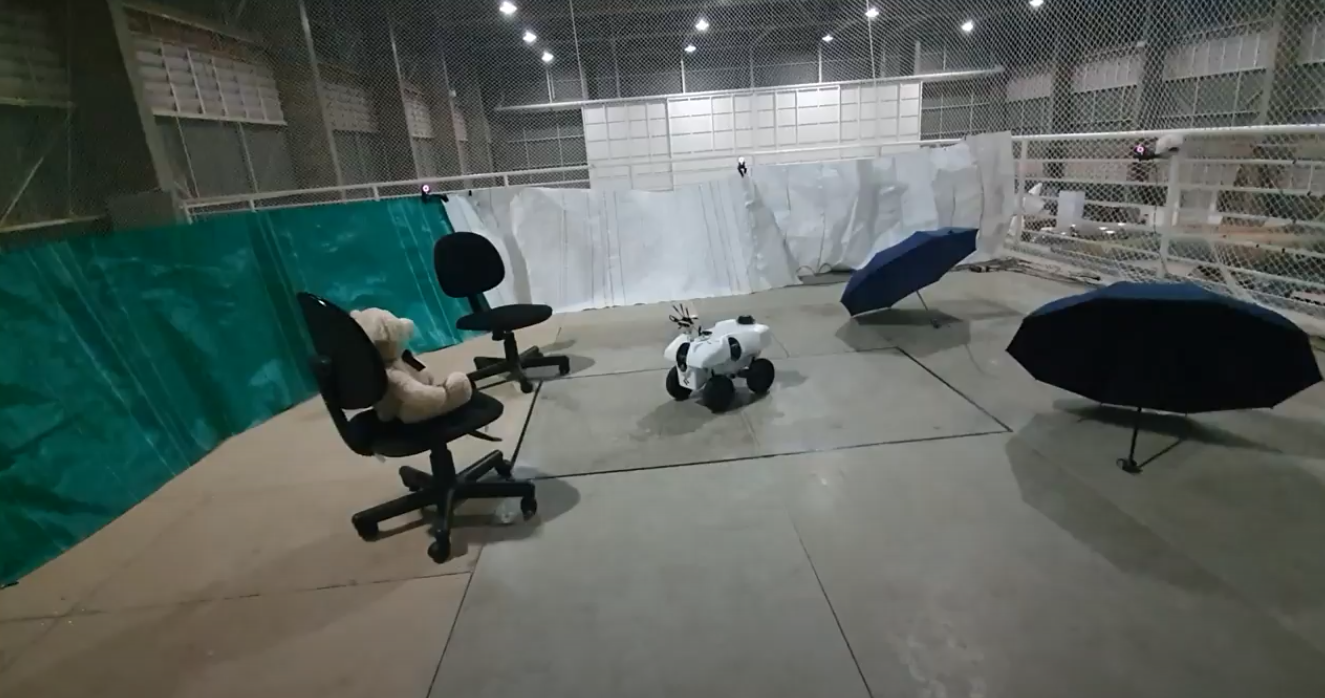} \label{fig:scene1_changing}
  } 
  \subfigure[       Scene 2]{%
    \includegraphics[width=.32\linewidth]{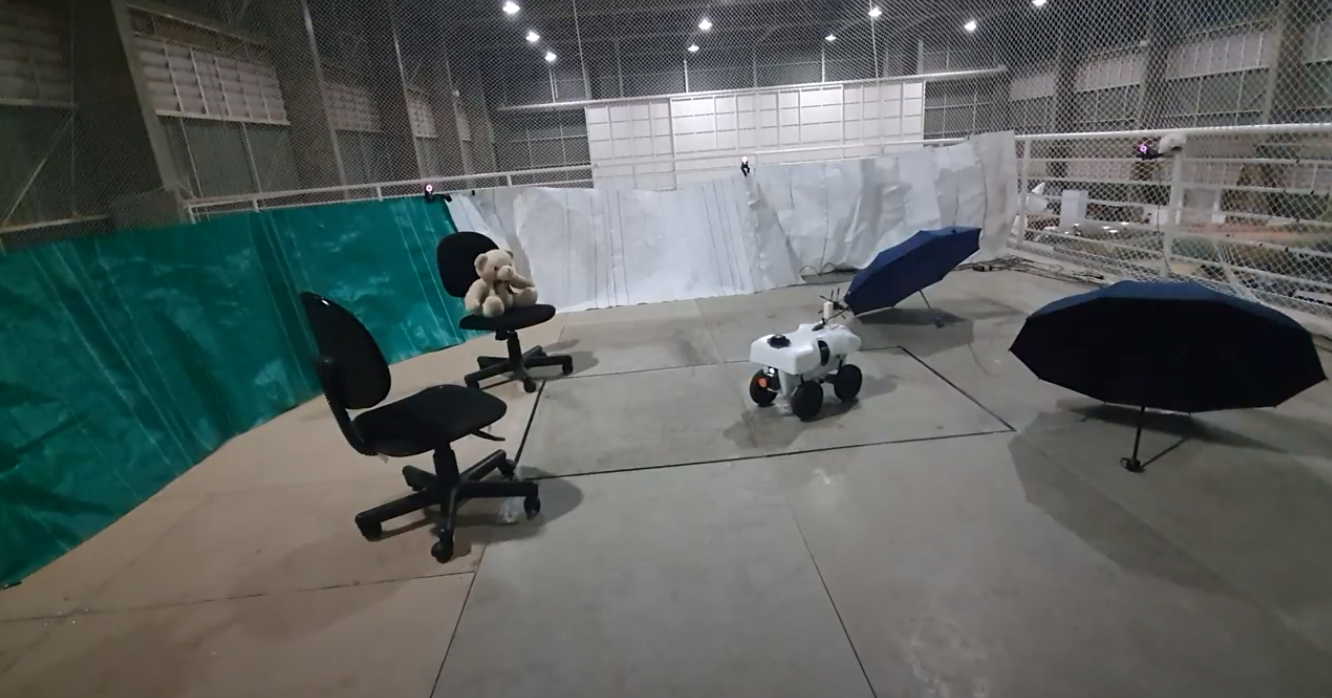} \label{fig:scene2_changing}
  } 
  \subfigure[        Scene 3]{%
    \includegraphics[width=.32\linewidth]{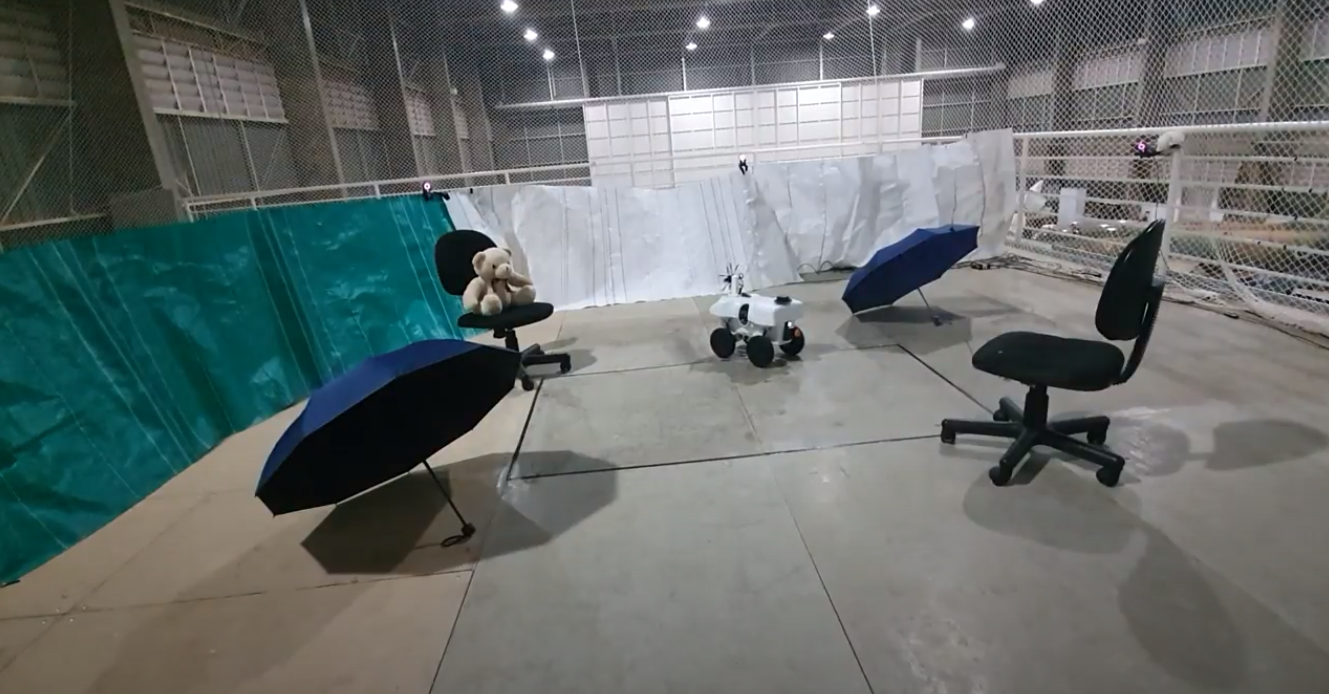}\label{fig:scene3_changing}
  } 
  \caption{Selected scenes from the \textit{Changing} sequence}
  \label{fig:changing_seq}
\end{figure*}

The \textit{Changing} sequence initially contains two umbrellas and two chairs. Also, there is one teddy bear on one of the chairs. It starts with the robot facing the chair with the teddy bear. As the robot wanders within the scene, the teddy bear is moved to the other chair. This can potentially trigger a wrong loop closure. Selected scenes from this sequence are shown in Fig.~\ref{fig:changing_seq}.

The \textit{Shift} sequence initially contains an umbrella on the floor, a teddy bear on a chair, and a suitcase with a monitor and a mug. The robot starts facing the chair with the teddy bear. As the robot wanders within the scene, the umbrella and chair are moved to other positions. After several turns, the suitcase together with the monitor and mug are also moved. This sequence is suitable to evaluate the ability of SLAM systems to detect changes of multiple objects.

The \textit{Replacing} sequence initially contains a chair with books, another chair with a teddy bear on it, and a suitcase with a monitor and a mug. The robot starts facing the chair with the teddy bear. After a while, the chair with the teddy bear is replaced by an umbrella. Finally, the other chair is removed.

\section{RESULTS}

\subsection{TUM Dataset}

Five dynamic sequences from the TUM Dataset were used to evaluate the robustness of the proposed method: the four walking sequences and the fr3\_sitting\_xyz sequence, to be used as a baseline. The system was compared to 14 systems from the literature. The comparison includes feature-based methods designed for static environments, such as ORB-SLAM3~\cite{orbslam3}, direct methods designed for dynamic environments, such as StaticFusion~\cite{static-fusion} and ReFusion~\cite{re-fusion}, feature-based methods designed for dynamic environments, for instance DynaSLAM~\cite{dynaslam}, DS-SLAM~\cite{dsslam}, SaD-SLAM~\cite{sadslam}, and DOTMask~\cite{dotmask}, and DX-SLAM \cite{dxslam}, a method for changing environments.

The results from Liu et al.~\cite{liu}, DynaSLAM, DS-SLAM, SOF-SLAM~\cite{sof-slam}, DetectSLAM~\cite{detect-slam}, SaD-SLAM, DOTMask, Ji et al.~\cite{ji2021} were obtained in their respective publications. The results from ORB-SLAM3 were obtained running the code using 1500 keypoints in each frame. Finally, the results from StaticFusion and ReFusion were obtained in~\cite{re-fusion}. For the results from the proposed work, the tests were performed 5 times and the median results were used for the evaluation, as proposed by Mur-Artal and Tardós~\cite{orbslam}.

\begin{figure}[ht] 
  \subfigure[      ORB-SLAM3]{%
    \includegraphics[width=.480\linewidth]{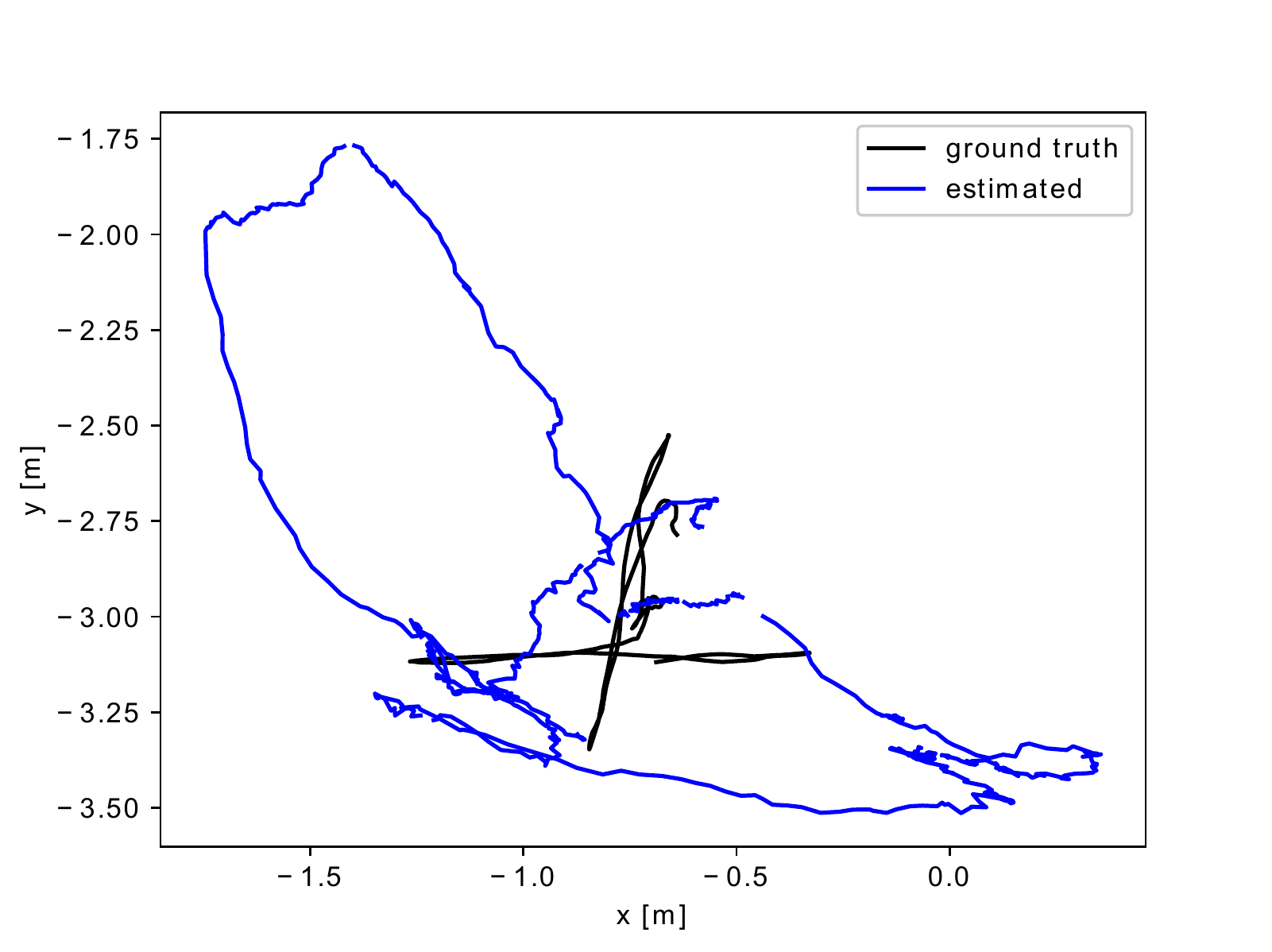} \label{fig:fr3wxyz_orbslam3_changingchapter}
  } 
  \subfigure[       Changing-SLAM]{%
    \includegraphics[width=.480\linewidth]{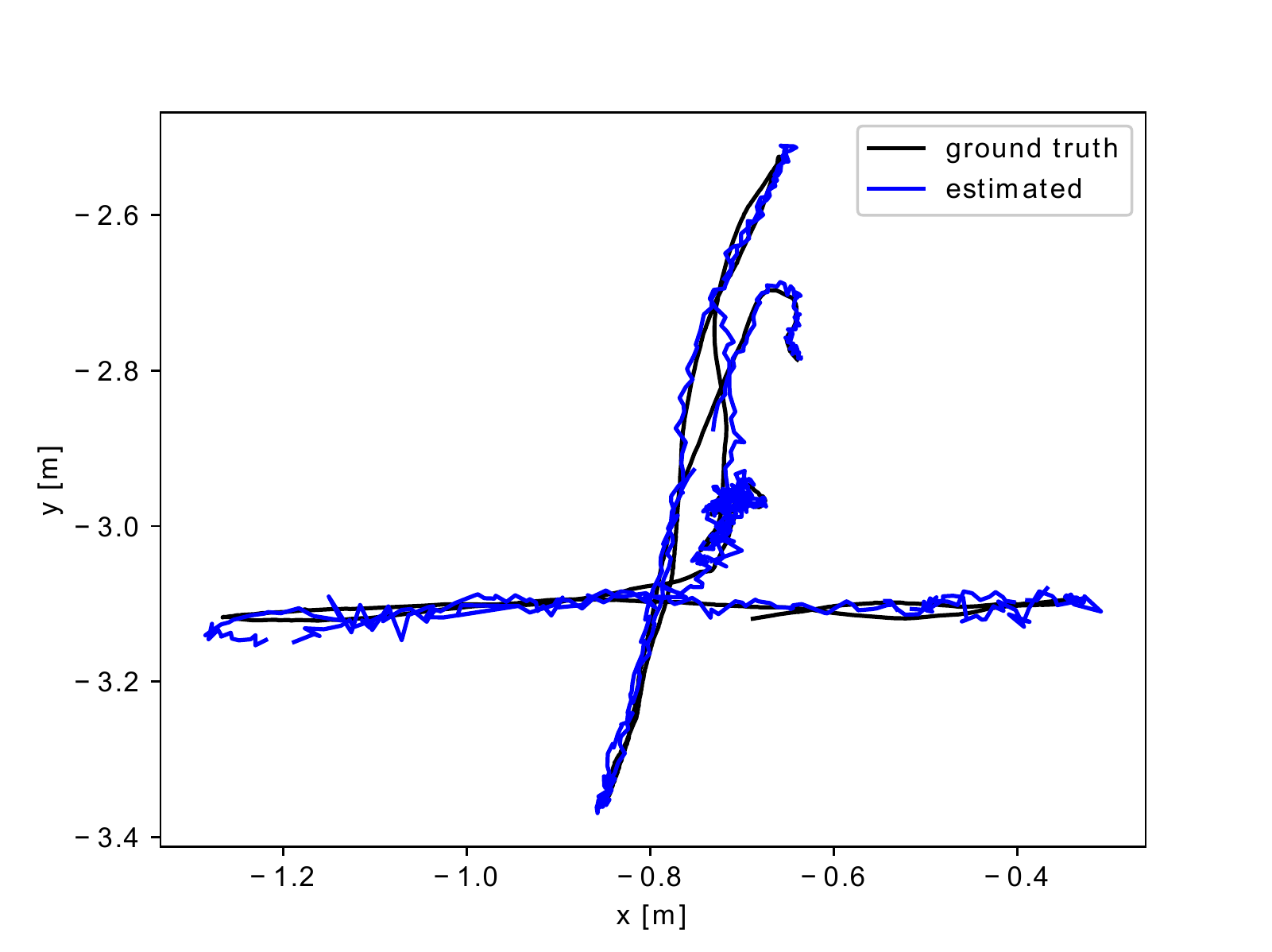} \label{fig:fr3wxyz_changingslam}
  } 
  \caption{Ground truth and estimated trajectory in the sequence fr3\_w\_xyz}
\end{figure}

Figures \ref{fig:fr3wxyz_orbslam3_changingchapter} and \ref{fig:fr3wxyz_changingslam} show, respectively, the comparison between the ground-truth and estimated trajectories of ORB-SLAM3 and Changing-SLAM for the fr3\_w\_xyz sequence. The trajectory estimated by ORB-SLAM3 completely deviates from the ground-truth.

Table \ref{ate_comparison_chapterdyn} shows the RMSE of the ATE comparison between the proposed method and the 14 methods. The bold values represent the best results in each sequence, and the underlined values represent the second best ones. The results show that ORB-SLAM3 cannot cope with dynamic environments. Their results are satisfactory only in the sitting sequence, were the people in the scene are sitting, just moving their hands.

DynaSLAM and SaD-SLAM had, overall, the best results. However, both methods are not real-time. SaD-SLAM, for instance, can only work offline. Our method achieved similar results working in real time. Changing-SLAM also outperformed all direct methods and two real-time feature-based ones: DOTMask and Detect-SLAM. The results also show that DX-SLAM is not robust to dynamic environments.

Overall, Changing-SLAM achieved the second best result in the challenging fr3\_w\_xyz sequence, and achieved similar results in the other sequences in comparison to the other systems, showing that Changing-SLAM is robust to dynamic environments.

\begin{table}[h]
\caption{Comparison of the RMSE of ATE [m] of the proposed method against ORB-SLAM3, StaticFusion, ReFusion, DynaSLAM, DS-SLAM, SOF-SLAM, Detect-SLAM, Liu \textit{et al.}, Sun \textit{et al.}, Sun \textit{et al.}, SaD-SLAM, DOTMask, Ji \textit{et al.}, and DX-SLAM using the TUM dataset}
\label{ate_comparison_chapterdyn}
\begin{center}
\setlength{\tabcolsep}{1.2pt}
\renewcommand{\arraystretch}{1.4}
\begin{tabular}{|c|c|c|c|c|c|c|c|}
\hline
Sequence & \textit{fr3\_s\_xyz} & \textit{fr3\_w\_static} & \textit{fr3\_w\_xyz} & \textit{fr3\_w\_rpy}& \textit{fr3\_w\_half} \\
\hline
Changing-SLAM  & 0.018 & 0.008 & \underline{0.016} & 0.067  &  0.039 \\
\hline
ORB-SLAM3 & \textbf{0.009}  & 0.038  & 0.819 & 0.957  &  0.315 \\
\hline
StaticFusion & 0.039   &  0.015 & 0.093 & --- & 0.681 \\
\hline
ReFusion &  0.040   &  0.017 & 0.099 & --- & 0.104 \\
\hline
DynaSLAM & 0.015   &  \textbf{0.006} & \textbf{0.015} & 0.035 & \textbf{0.025} \\
\hline
DS-SLAM & ---  & 0.008  & 0.024 & 0.444 &  0.030 \\
\hline
SOF-SLAM   & ---  & \underline{0.007}  & 0.018  &  \textbf{0.027} & 0.029 \\
\hline
Detect-SLAM   & 0.020  & ---  & 0.024  &  0.296 & 0.051 \\
\hline
Liu \textit{et al.} \cite{liu} & ---  & 0.011  & \underline{0.016}  &  0.042 & 0.031 \\   
\hline
Sun \textit{et al.} \cite{sun2017}  & 0.048  & 0.065  & 0.093  &  0.133 & 0.125 \\
\hline
Sun \textit{et al.} \cite{sun2018}  & 0.051  & 0.033  & 0.066  &  0.073 & 0.067 \\
\hline
SaD-SLAM    & \underline{0.012} & 0.017 & 0.017  & \underline{0.032}  &  \underline{0.026}  \\
\hline
DOTMask   & 0.018 & 0.008 & 0.021  & 0.053  &  0.040 \\
\hline
Ji \textit{et al.} \cite{ji2021}  & \underline{0.012} & 0.011 & 0.020  & 0.037  & 0.029\\
\hline
DX-SLAM  & --- & 0.017 & 0.309  & ---  & ---\\
\hline
\end{tabular}
\end{center}
\end{table}

\subsection{PUC-USP Dataset}

The PUC-USP Dataset, presented in Section~\ref{sec: puc-usp-dataset}, was used to evaluate the robustness of the proposed methodology in changing environments.

\begin{figure*}[h] 
  \subfigure[  \hspace{0.05cm}    Feature detection ]{%
    \includegraphics[width=.31\textwidth]{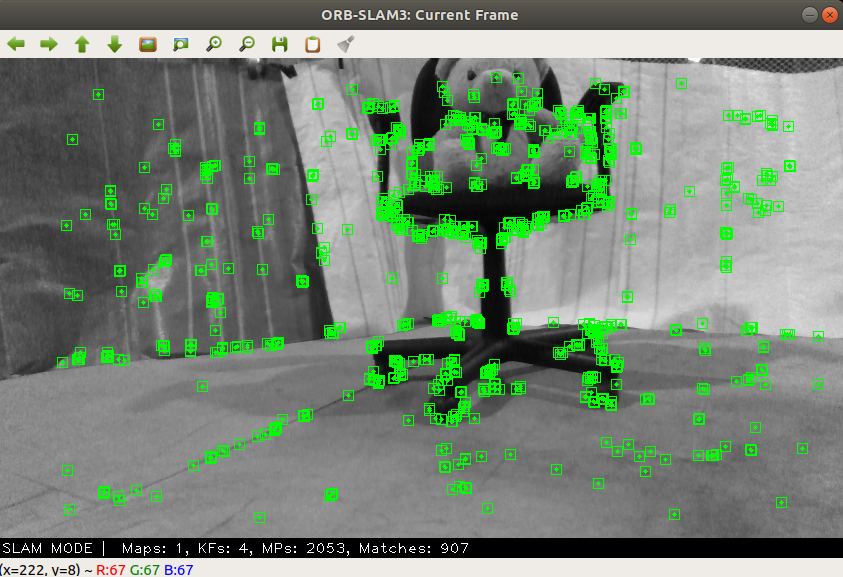}
    \label{fig:onechair_featuredetection} 
  } 
  \subfigure[ \hspace{0.05cm}   Initial classification]{%
    \includegraphics[width=.31\textwidth]{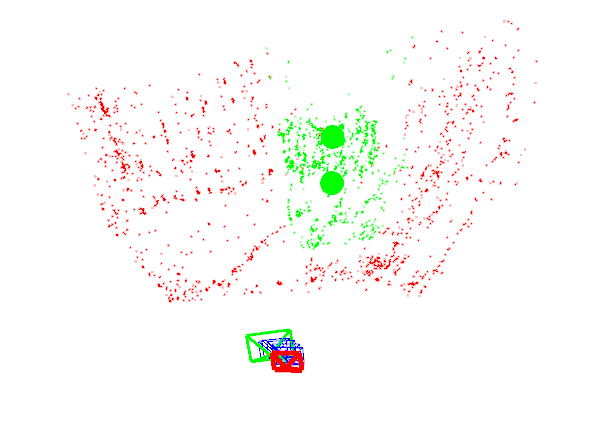} \label{fig:onechair_initialmappoints} 
  }
  \subfigure[ \hspace{0.05cm}   Filtered map]{%
    \includegraphics[width=.31\textwidth]{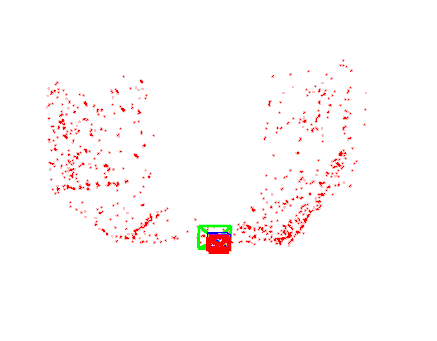} \label{fig:onechair_filtered} 
  }     
  \caption{MapPoint filtering process performed by Changing-SLAM in the \textit{OneChair} sequence. The two larger green dots represent two MapObjects in the map. The small green dots in the map represent the MapPoints associated to the MapObjects.} 
\end{figure*}

Figures \ref{fig:onechair_featuredetection}-\ref{fig:onechair_filtered} show an example of the proposed method in the \textit{OneChair} sequence. There are two objects in the scene: a chair and a teddy bear. First, the features are detected in the RGB image, as shown in Fig. \ref{fig:onechair_featuredetection}. The keypoints are classified as belonging to the objects or to the background. Figure \ref{fig:onechair_initialmappoints} shows the MapPoints in green classified as belonging to the objects, and the MapPoints in red as from the background. The larger green dots represent the centroids of the MapObjects. The green MapPoints are initially removed from the map, as shown in Fig. \ref{fig:onechair_filtered}, as they belong to MapObjects which beliefs are 0.5. However, both objects are detected, estimated and stored in memory. If the objects are observed again later, their belief would increase and they would be inserted again in the map together with their MapPoints.

\begin{figure}[ht] 
  \subfigure[        ORB-SLAM3]{%
    \includegraphics[width=.480\linewidth]{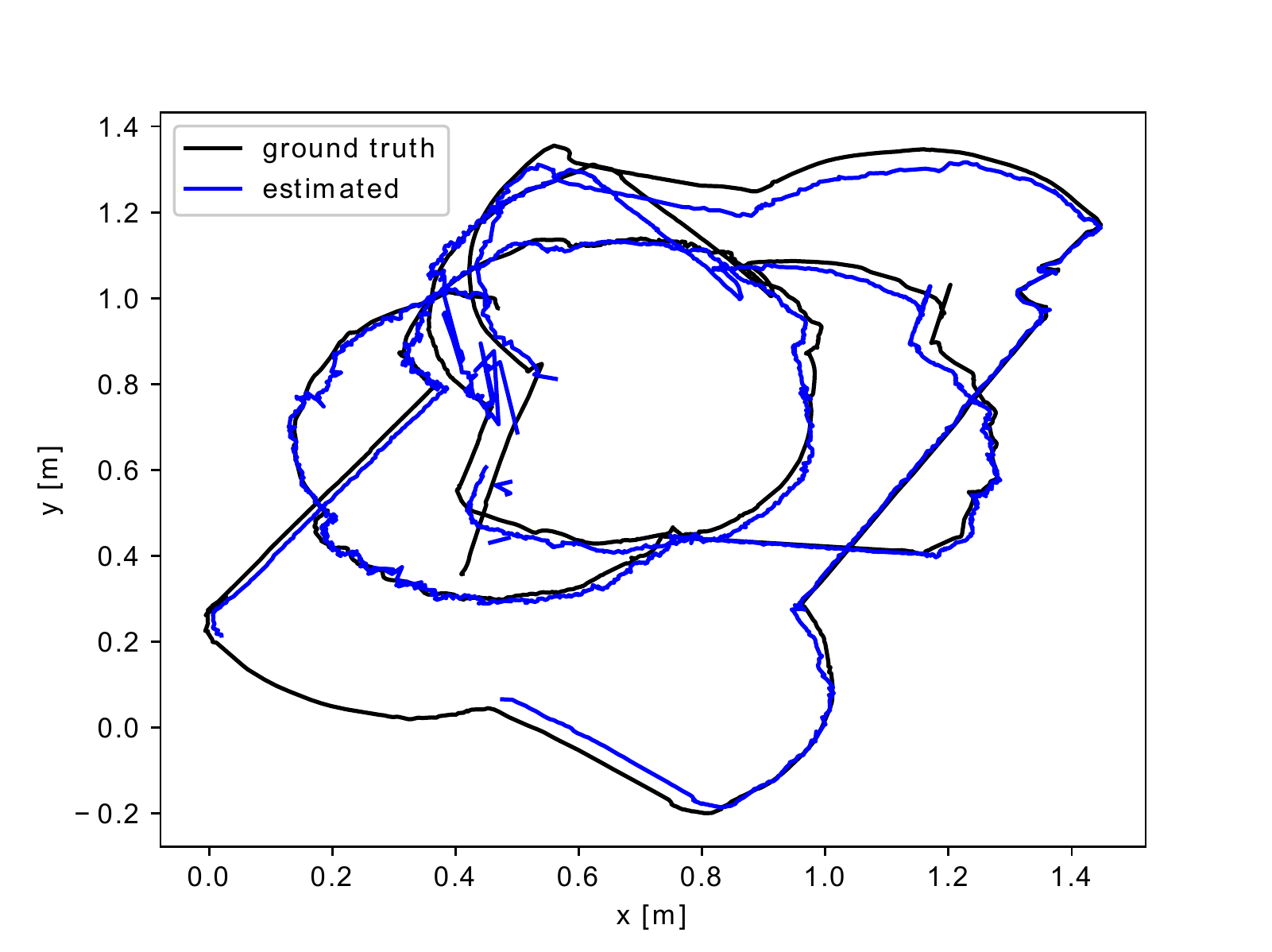} \label{fig:changing_orbslam3}
  } 
  \subfigure[        Changing-SLAM]{%
    \includegraphics[width=.480\linewidth]{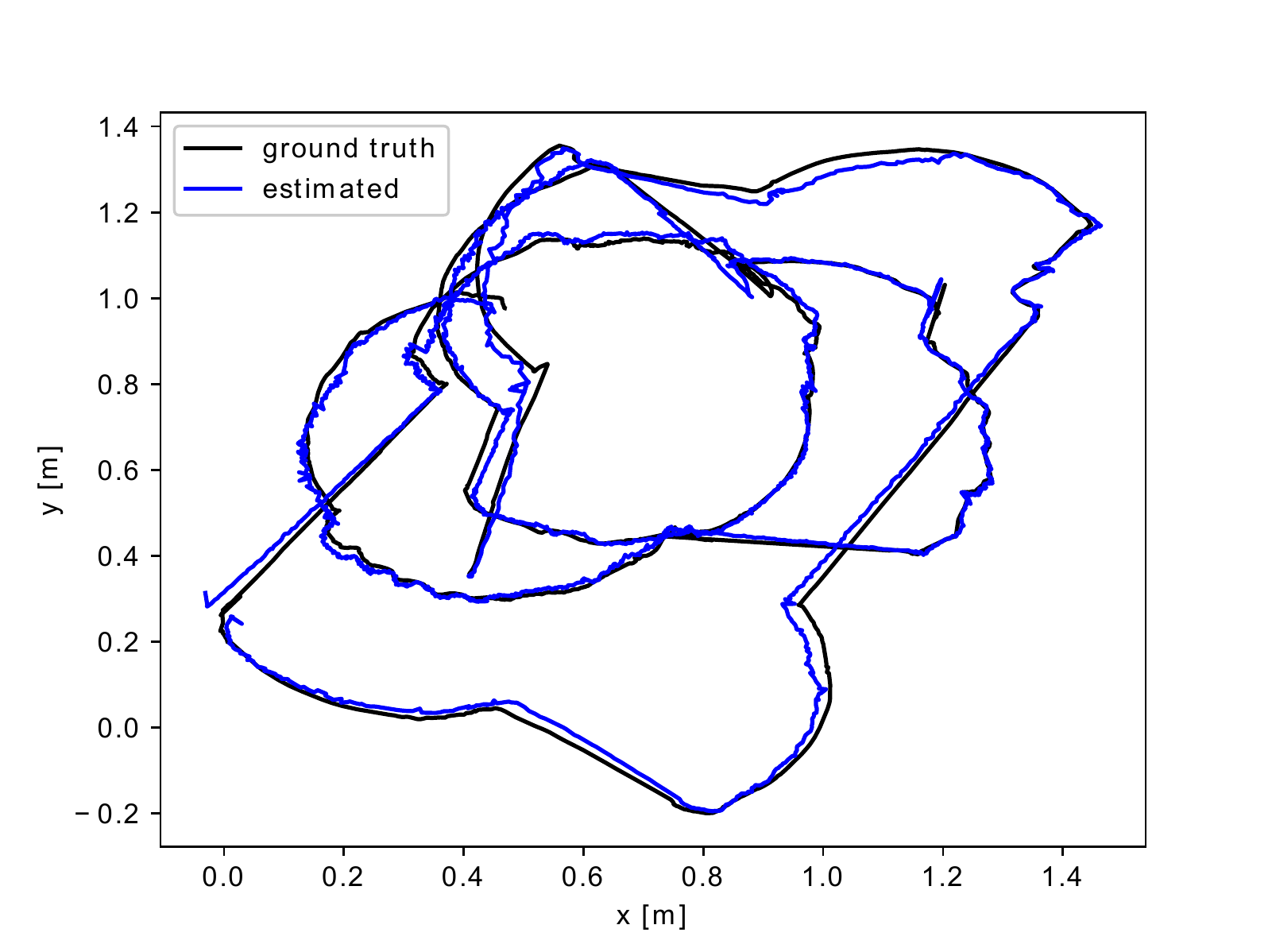}
    \label{fig:changing_changingslam}
  } 
  \caption{Comparison of ground truth and estimated trajectory in the sequence \textit{Changing}} 
\end{figure}

Table \ref{ate_comparison_pucusp_changing} shows the ATE comparison between ORB-SLAM3, DXSLAM and Changing-SLAM. As expected, all systems achieved good results in the static scenario. Despite the fact that Changing-SLAM improved every result of ORB-SLAM3, it is noticeable that the effect of changing environments is not always critical for the localization accuracy, as it happens in dynamic environments. The $Vanishing$ sequence was not expected to cause a major error, because there are no new objects added in the scene, which could have caused a wrong loop closure. The $Changing$ sequence caused an increase in the localization error of ORB-SLAM3, as shown in Figs. \ref{fig:changing_orbslam3} and \ref{fig:changing_changingslam}. ORB-SLAM3 had lost tracks and drifts that were corrected by Changing-SLAM. The sequence \textit{OneChair} triggered a wrong loop closure in ORB-SLAM3, which caused a considerable increase in the RMSE. 


\begin{figure}[ht] 
  \subfigure[   \hspace{0.05cm}     ORB-SLAM3]{%
    \includegraphics[width=.48\linewidth]{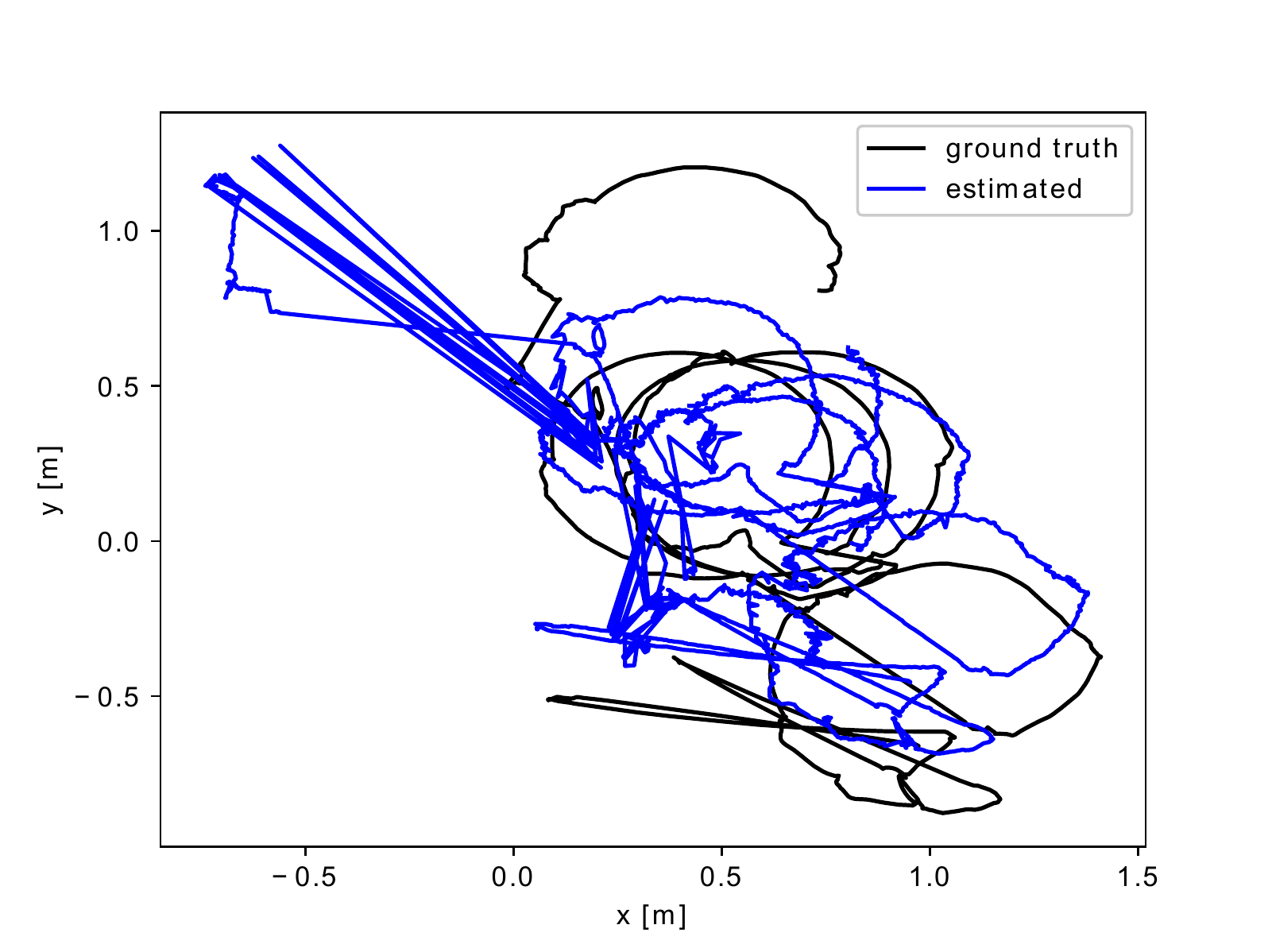} \label{fig:onechair_orbslam3}
  } 
  \subfigure[   \hspace{0.05cm}     Changing-SLAM]{%
    \includegraphics[width=.48\linewidth]{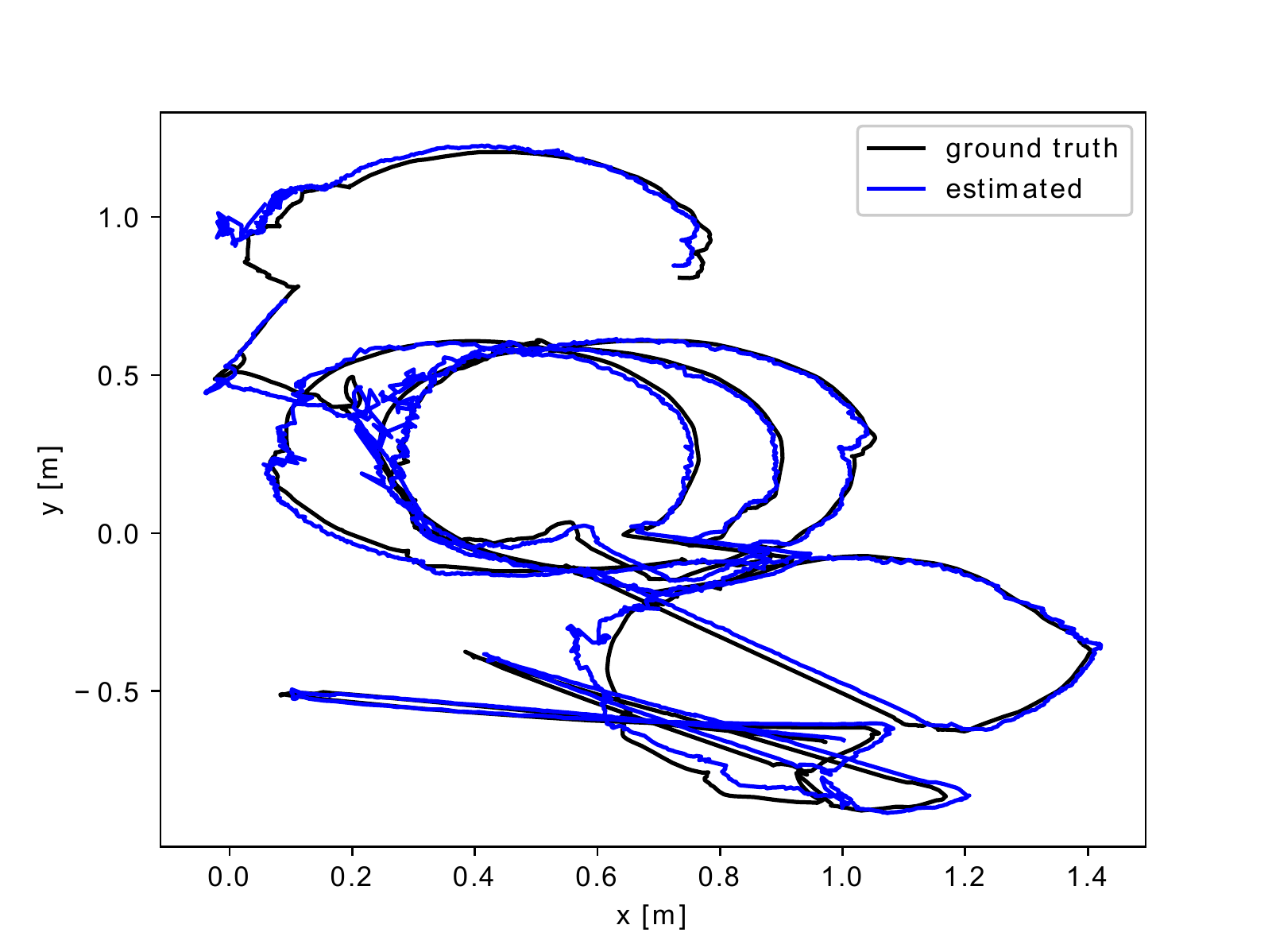} \label{fig:onechair_changing}
  } 
  \caption{Comparison of ground truth and estimated trajectory in the sequence \textit{OneChair}} 
\end{figure}

\begin{figure}[ht] 
  \subfigure[   \hspace{0.05cm}     Before loop closure]{%
    \includegraphics[width=.49\linewidth]{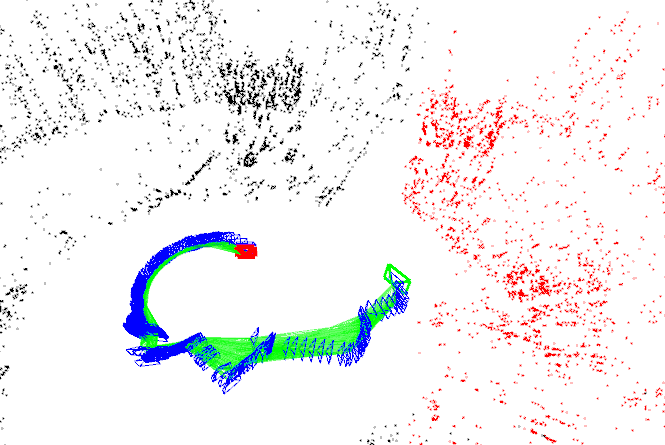} \label{fig:wrongloop1_orb3_onechair}
  } 
  \subfigure[   \hspace{0.05cm}     After loop closure]{%
    \includegraphics[width=.44\linewidth]{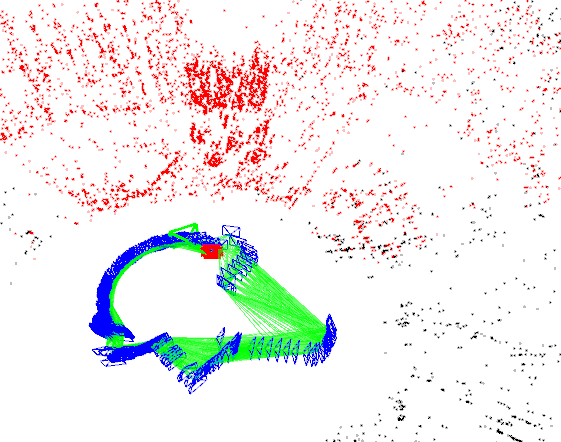} \label{fig:wrongloop2_orb3_onechair}
  } 
  \caption{Wrong loop detection by ORB-SLAM3 in the OneChair sequence}
\end{figure}

Figures \ref{fig:wrongloop1_orb3_onechair} and \ref{fig:wrongloop2_orb3_onechair} show the wrong loop closure made by ORB-SLAM 3 in the \textit{OneChair} sequence. The shapes of two chairs are noticeable in Fig. \ref{fig:wrongloop1_orb3_onechair}, merged into one chair in Fig. \ref{fig:wrongloop2_orb3_onechair}. The wrong loop occurred due to the change in the chair position and the inability of ORB-SLAM3 to detect this change. After this wrong loop closure, the system is no longer able to recover from the error, even with the robust multi-mapping and re-localization systems of ORB-SLAM3, as shown in Fig. \ref{fig:onechair_orbslam3}.

\begin{table}[h!]
\caption{Comparison of the RMSE of ATE [m] of Changing-SLAM against ORB-SLAM3 and DX-SLAM using the PUC-USP dataset}
\label{ate_comparison_pucusp_changing}
\begin{center}
\renewcommand{\arraystretch}{1.2}
\begin{tabular}{|c|c|c|c|c|c|}
\hline
Sequence & ORB-SLAM3 & DX-SLAM & Changing-SLAM\\
\hline
\textit{Static} &  \textbf{0.033} & 0.036 &  \textbf{0.033}\\
\hline
\textit{OneChair} & 0.407 & 0.097  & \textbf{0.089}\\
\hline
\textit{Vanishing} & 0.052  & 0.062 & \textbf{0.049} \\
\hline
\textit{Changing} & 0.071  & 0.044 &  \textbf{0.029} \\
\hline
\textit{Shift} & \textbf{0.075}  & 0.077 & \textbf{0.075} \\
\hline
\textit{Replacing} & 0.049 & 0.055 & \textbf{0.047} \\
\hline
\end{tabular}
\end{center}
\end{table}

\subsection{Run-time Analysis}

All tests were performed on a notebook with an Intel Core i7-10750H and 16 GB of RAM running Ubuntu Linux 18.04 LTS. The system is implemented in C++, and the object detection is performed with OpenCV 4.5, using a Nvidia GeForce RTX 2060 GPU. Changing-SLAM achieved an average tracking speed of 23.8 FPS, considering all steps including object detection, which can be categorized as real time. Table \ref{tab:frame_rate_changing} compares the tracking speed of Changing-SLAM with other state-of-the-art systems. DOTMask \cite{dotmask} achieves 14.3 FPS using a GTX 1080 GPU in the object tracking pipeline, without considering the visual SLAM processing time. Ji et al. \cite{ji2021} achieves a tracking speed of 13.2 FPS without considering the instance segmentation.

\begin{table}[ht]
\caption{Mean tracking speed [FPS] comparison}
\label{tab:frame_rate_changing}
\begin{center}
\renewcommand{\arraystretch}{1.2}
\begin{tabular}{|c|c|}
\hline
System &  Mean Frame Rate [FPS]\\
\hline
Changing-SLAM &  23.8\\
\hline
DOTMask &  14.3\\
\hline
Ji et al.  &   13.2\\
\hline
DynaSLAM &   2.1\\
\hline
\end{tabular}
\end{center}
\end{table}

\subsection{Limitations}

Changing-SLAM does not deal with deformable objects such as blankets, rigid objects that can change shape such as cabinets with closed and open doors, and objects not labeled in the object detector training process. The latter problem can be solved re-training the network to include more objects that are common to the environment. 

The COCO Dataset \cite{coco} has several classes that are not suitable for indoor environments, which is the scope of this work, such as cars, and even other classes that are not common for regular outdoor situations, such as giraffes and other wild animals. Therefore, it would be beneficial to train the network with more common indoor objects. However, this approach still cannot deal with unexpected new objects, as performed in the approach proposed by Ji et al. \cite{ji2021}.

\section{CONCLUSIONS}

This work presented Changing-SLAM, our proposed method to perform visual SLAM in both dynamic and changing scenarios in real time. To our knowledge, this is the first system able to perform these tasks simultaneously in real time only using a camera.

The proposed method was tested with a dataset especially designed for the evaluation of visual SLAM systems in changing scenarios, achieving a high accuracy in comparison with ORB-SLAM3 and DXSLAM. Changing-SLAM is very robust in such scenarios, preventing the detection of a wrong loop closure that would ruin the SLAM process.

Besides correcting localization, the semantic map generated by Changing-SLAM can be useful for a wide variety of applications. One example would be within the work from Chen and Liu \cite{spaceconstruction}, which generates navigable paths from the maps generated by ORB-SLAM2 and ORB-SLAM3. These maps would be corrupted if objects were moved in the scene. 

Finally, the use of object detection and feature repopulation to differentiate object features from the background ones is a method to decrease the computational effort of the system. However, the semantic mapping, dynamic object filtering and belief update methods are not restricted to that. The proposed method can be performed using other types of semantic detection such as instance or panoptic segmentation, when they become computationally feasible. Therefore, for future work, panoptic segmentation could be used in the methodology to evaluate if there is an improvement in accuracy.

\section*{Acknowledgments}
This research  was financed in part by the Coordenação de Aperfeiçoamento de Pessoal de Nível Superior - Brasil (CAPES) - Finance Code 001, and in part by FAPESP, process number 2021/05336-3.

\bibliographystyle{IEEEtran}
\bibliography{main}

\vspace{11pt}

\vfill

\end{document}